\documentclass[11pt]{article}
\usepackage{UF_FRED_paper_style}
\usepackage{algorithm}
\usepackage{algpseudocode}
\usepackage{commath}
\usepackage{lipsum}  
\usepackage{mathtools}
\usepackage{amsfonts}
\usepackage{calrsfs}
\usepackage{romannum}
\usepackage{xcolor}
\DeclareMathAlphabet{\pazocal}{OMS}{zplm}{m}{n}

\newcommand{\Xmat}{\mathbf{X}}
\newcommand{\Ymat}{\mathbf{Y}}
\newcommand{\Amat}{\mathbf{A}}
\newcommand{\Xset}{\mathbb{X}}

\newcommand{\graph}{\mathcal{G}}

\newcommand{\ones}{\mathbf{1}}

\title{Population-wise Labeling of Sulcal Graphs using Multi-graph Matching}

\author{R.Yadav$^{\star\ddag\dagger}$, F.X. Dup\'e$^{\dagger}$, S. Takerkart$^{\star}$, G. Auzias$^{\star}$\\
$^{\star}$Institut de Neurosciences de la Timone UMR 7289, Aix-Marseille Universit\'e, CNRS \\
$^{\ddag}$Aix Marseille Universit\'e, Institut Marseille Imaging, Marseille, France\\
$^{\dagger}$Laboratoire d'Informatique et Syst\`emes UMR 7020, Aix-Marseille Universit\'e, CNRS
}

\date{\today}

\begin{document}
\maketitle


\begin{abstract}

Population-wise matching of the cortical fold is necessary to identify biomarkers of neurological or psychiatric disorders. 
The difficulty comes from the massive inter-individual variations in the morphology and spatial organization of the folds.
This task is challenging at both methodological and conceptual levels.
In the widely used registration-based techniques, these variations are considered as noise and the matching of folds is only implicit. 
Alternative approaches are based on the extraction and explicit identification of the cortical folds.
In particular, representing cortical folding patterns as graphs of sulcal basins -- termed \textit{sulcal graphs} -- enables to formalize the task as a graph-matching problem.
In this paper, we propose to address the problem of sulcal graph matching directly at the population level using multi-graph matching techniques.
First, we motivate the relevance of multi-graph matching framework in this context.
We then introduce a procedure to generate populations of artificial sulcal graphs, which
allows us benchmarking several state of the art multi-graph matching methods. 
Our results on both artificial and real data demonstrate the effectiveness of multi-graph matching techniques to obtain a population-wise consistent labeling of cortical folds at the sulcal basins level.

\end{abstract}


\section{Introduction}
\subsection{Quantitative comparison across brains is a crucial but open question}

Comparing features extracted from brain MRI across individuals is necessary for estimating population statistics and ultimately discover markers of diseases.
However, this task presents several challenges at both the methodological and conceptual levels. 
Indeed, the features extracted from two different individual brains are defined in two different mathematical spaces. Comparing such features thus requires to address the \textit{methodological} problem of transferring them into a common space. 
The task of transferring information from one brain to another or to a common space consists in defining spatial correspondences across these objects by compensating for their variations in their respective geometry.
The challenge lies in the massive inter-individual variations of the morphology of the brain and in particular the geometry of the cortical surface, which make the identification of such spatial correspondences an ill-posed problem.
As a consequence, any solution to this problem inevitably requires to introduce additional constraints based on assumptions on the biological validity of the resulting spatial correspondences, which constitutes a challenge at the \textit{conceptual} level.
Indeed, the assumptions and constraints introduced in the definition of the spatial correspondences actually influence the derived statistics measured on the population of interest, and could thus be considered as a source of bias in the analysis \cite{van_essen_parcellations_2012}.

One widely used approach to tackle this problem -- termed here as the registration-based approach -- consists in defining a mapping between each individual brain and an atlas serving as the common space by estimating a spatial transformation. As pointed above, the process of building the atlas and defining the associated projection operator which minimizes the error induced by the transformation remains an open research question. As a consequence, several registration techniques and atlases co-exist in the field, and tools to enable comparison \textit{across atlases} are then required \citep{devlin_praise_2007,van_essen_navigating_2007}.
The variety of atlases, projection mechanisms and descriptors illustrate the ongoing exploration of putative biologically relevant features used to define these correspondences across individuals.
One of the most widely used registration-based approach \citep{fischl1999high} defines a mapping between cortical surfaces by imposing the alignment of a combination of curvature and convexity features estimated from a 2D mesh representing the geometry of the cortex.
The cortical surface of a given subject is projected onto the atlas by matching its curvature and convexity, under the assumption that aligning these features induces biologically relevant anatomo-functional correspondences. 
In this process, as in any registration-based approach, variations across individuals are considered as noise or confounding perturbations to be minimized, including variations in the topology and number of folds (sulci).
More generally, the registration-based approach might be seen as an over simplification of the problem since potentially relevant geometrical information is not taken into account.

Alternative approaches consist in characterizing the geometry and organization of the cortical folds in each individual and then compare these features across the population.

\subsection{Characterizing cortical folding patterns using graphs}
\label{sec:bib_sulcal_graphs}
\begin{figure}[htb]
    \centering
    \includegraphics[width=0.9\linewidth]{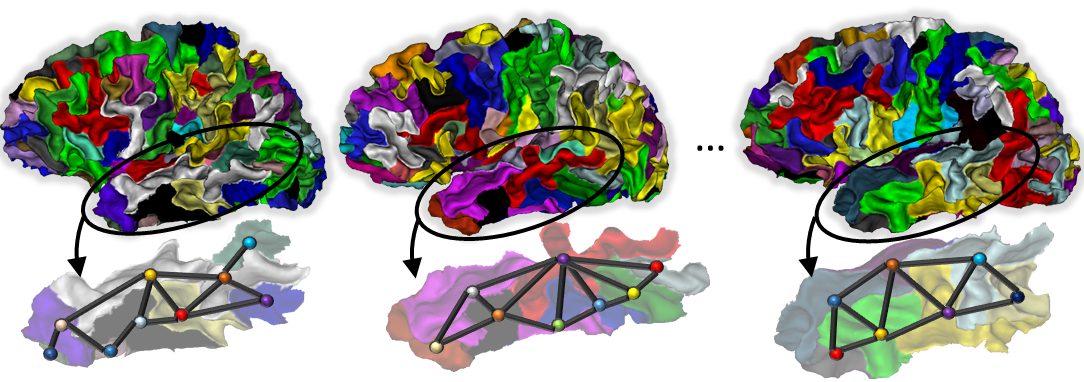}
    \caption{Example of sulcal graphs from three individual brains, superimposed with the underlying decomposition of the cortical surface in sulcal basins. Sulcal basins are shown in different colours, and their corresponding node in the graph are represented as spherical dots in the lower panel. The color of each node in the graph illustrates the value of a given attribute such as for instance the area or depth of corresponding sulcal basin.}
    \label{fig:example_sulcal_pits_graphs}
\end{figure}

Several approaches have been proposed to characterize cortical folding patterns, such as gyrification index, fractal dimension and  curvature \citep{armstrong1995ontogeny, cachia2008cortical,im2006fractal}.
Although these measures capture relevant morphological features, they do not explicitly reflect the topology, i.e the spatial relationships between sulci.
\citet{mangin2004object} introduced an analysis framework based on the automatic extraction and labeling of the sulci allowing to characterize their shape, size and pattern in terms of e.g. sulcus area, depth and length. This representation of the cortical geometry has been used for instance to characterize populations of healthy subjects \citep{duchesnay_classification_2007}, to quantify potential deviations from normal populations in various conditions such as schizophrenia \citep{cachia2008cortical} and autism spectrum disorder \citep{auzias2014atypical}, or to estimate the heritability of the folding patterns \citep{pizzagalli_reliability_2020}.
Pursuing on this line of research, the \textit{sulcal pits} were introduced as a concept allowing to decompose the sulci into smaller pieces and thus access finer scale geometrical information.
As described in details in \citet{im_spatial_2010,auzias_deep_2015}, each fold is divided into \textit{sulcal basins} that are defined as concavities in the white matter surface bounded by convex ridges, and the deepest point in each basin defines the associated sulcal pit.
More recently, \citet{im2011quantitative, takerkart_structural_2017} represented the geometrical relationships between sulcal basins as a \textit{sulcal graph}.
A sulcal graph is constructed by considering each sulcal basin (or associated pit) as a node, while the edges connect only adjacent basins and thus represent their spatial organization.
Various geometrical features of a sulcal basin can then be attributed to graph nodes (such as the depth of the pit, its 3d coordinates...), while the spatial organization of the basins is encoded in the topology of the graph.
Figure~\ref{fig:example_sulcal_pits_graphs} illustrates this decomposition of the cortical folds into sulcal basins allowing to represent this complex geometry as a sulcal graph.

These sulcal graphs constitute particularly relevant representations because:
1) variations across individuals are preserved and are manifested as changes in both the topology of the graph and the value of the attributes attached to the nodes and edges; 2) the design of tools for the quantitative characterization of these variations can benefit from the extensive body of methods from the graph processing literature.

\subsection{Problem statement and contributions}

In the present work, we focus on the task of \textit{matching together a set of sulcal graphs} in order to define biologically relevant \textit{correspondences across a population of subjects}, under the specific constraint of explicitly taking into account the variations in \textit{folding patterns}. 
Before moving to the formalization, we more precisely situate this problem with respect to the conceptual question of defining correspondences across individuals, and with respect to the methodological problem of graph matching.

\subsubsection{Unsupervised comparison and matching of sulcal graphs}
\label{sec:abs_GT}

Comparing brains using sulcal graphs is highly relevant because all the geometrical information about the macroscopic cortical folding can be encoded into such graphs.
However, several challenges need to be addressed in this context: 
1) the large inter-individual variations in brain anatomy induce complex variations across sulcal graphs, including in their topology; 
2) sulcal graphs can be contaminated by noise resulting from the imperfect segmentation of the individual cortical surface and corresponding sulcal basins; 
3) there is no consensus on a nomenclature or atlas at the scale of sulcal basins covering the whole brain, that is a prerequisite to tackle the matching problem as a supervised learning task.
Indeed, few studies investigated the matching of cortical folds across individuals as a supervised task \citep{riviere_automatic_2002,borne2020automatic, behnke_automatic_2003}. All these works focused at the scale of sulci, i.e. considering large folds consisting of several of our sulcal basins. 
To our knowledge, only \cite{lyu_labeling_2021} attempted to tackle this problem at finer scale, probably because of the massive amount of efforts needed to gather sufficient amount of manually labeled data \cite{voorhies_cognitive_2021}. 
Indeed, ambiguities due to variations across individuals in the folding patterns become overwhelming at finer scale than sulci.
This is illustrated by the tedious works advancing the definition of a fined-grained nomenclature of folds \cite{sprung-much_morphology_2020} and their relationship with underlying function \cite{willbrand_uncovering_2022}.
The lack of widely accepted fined-grained nomenclature is also blatant in the related field of brain parcellation: more than 20 different fine-grained atlases co-exist \citep{eickhoff_imaging-based_2018}, and even the most advanced multi-modal atlas \citep{glasser_multi-modal_2016} was validated only on a small portion of the cortex. 

Matching sulcal graphs across individuals is thus a very challenging problem.
Instead of relying on the few existing labeled data-sets that clearly deserve further validation, we decided to approach this question as an unsupervised learning task.

We now describe the few studies that have attempted to tackle the question of unsupervised labeling of sulcal graphs. The first approach was proposed by \citet{im_spatial_2010} and consisted in computing a map of the spatial density of sulcal pits across a population of subjects. This density map was computed by accumulating the pits from the different individuals in each vertex of an average surface after aligning the folds using a registration technique.
A watershed algorithm was then applied to this density map in order to separate the main \textit{clusters of sulcal pits}, empirically defined as the regions of high density. 
An arbitrary label was then associated to each cluster, hereby defining an \textit{ad-hoc} labeling of the pits across individuals, depending on the cluster to which they contributed in the density map.
This procedure implicitly defines a matching of sulcal pits and corresponding basins across individuals.
Exemplar applications of this method can be found in e.g. \cite{im_spatial_2010, auzias_deep_2015, le_guen_genetic_2017}, with illustrations of density maps and induced labeling for various populations.
We refer in the following to this category of methods as \textbf{Auzias et al.} since we used the open source implementation from this paper. 
The main limitation of this approach is that the labeling is driven only by the coordinates of the sulcal pit. 

\cite{irene_kaltenmark_cortical_2020} introduced an alternative procedure for labeling the sulcal basins, hereby considering the geometry of the basin surrounding each sulcal pit in addition to its spatial location.
We refer to this method as \textbf{Kaltenmark et al.} in the following.
The authors of \citep{irene_kaltenmark_cortical_2020} also raised the question of the \textit{consistency} of the labeling, a notion that we will develop further below.
In this method, an explicit constraint is imposed to restrict the labeling to only one node per subject for each label.
In addition, the nodes for which the labeling is ambiguous -- i.e. for which several labels are equally plausible -- remain unlabelled, which is often denoted as \textit{partial matching} in the literature on graph processing.
Importantly, the spatial relationships between adjacent sulcal basins and pits are never taken into account in any of these methods, since the different pits/basins from each subject are considered independently.
In contrast, in the present work our aim is to exploit the spatial organization of the adjacent basins stored in the sulcal graph representation.

Few publications investigated the potential of graph matching in the context of sulcal graphs.
In \cite{im2011quantitative}, the spectral graph matching technique \citep{leordeanu} was applied to a set of 48 monozygotic twins, comparing a pair at a time. This study showed that the similarity of the sulcal graphs across pairs of twins are higher than for unrelated pairs, demonstrating the genetic influence on sulcal patterns, and the relevance of graph matching techniques in this context.
This approach was used in follow-up papers from the same group, e.g for comparing brain lobes in \cite{morton_abnormal_2019} or for matching individuals onto an atlas in \cite{im_quantitative_2017}.

In the work by \cite{meng2018discovering}, a population of 677 neonates was analyzed based on a sulcal graph comparison method similar to the one of \cite{im2011quantitative}.
The authors proposed to use different features of the sulcal pits such as the pit position, the pit depth, the basin area, the basin boundary and the pit local connectivity to construct different similarity matrices, one per feature, and merge them into a single one using a matrix fusion technique \citep{wang2014similarity}.
A clustering algorithm was then applied to the fused similarity matrix to identify sub-populations of sulcal graphs, associated to specific folding patterns in the central, cingulate and superior temporal regions.

Critically, all these previous studies relied only on \textit{pairwise} graph matching techniques.
Comparing pairs of graphs independently, in the presence of noise and large inter-individual variations, is clearly sub-optimal.

\subsubsection{Multi-graph matching: a relevant framework for population studies}
\label{sec:MGM_population}

Given the large variations across subjects and imperfect sulcal basins extraction, examining jointly a group of sulcal graphs is key to reveal meaningful information not accessible by considering only pairs of subjects. This is the translation to sulcal graphs of the basic idea behind general population studies, that allowed researchers to uncover some of the mechanisms underlying the anatomo-functional organization of the brain.
We follow this principle by investigating for the first time the potential of \textit{multi-graph} matching techniques in the context of sulcal graphs.
By considering several brains together, the geometrical information that is shared by the majority of individuals should help to regularize the matching problem and allow to identify putative noisy graph nodes in a more robust way than with pairwise matching.
The multi-graph matching framework has the potential to uncover population-wise invariant patterns in sulcal graphs  without imposing a priori, potentially biasing, assumptions.


\subsubsection{Contributions}

In our previous work \citep{Buskulic2021}, we introduced a framework to generate a set of synthetic sulcal graphs representative of a population, and used it to benchmark state of the art \textit{pairwise} matching techniques in the context of sulcal graphs. 
In \cite{Yadav2022}, we provided a proof of concept of the relevance of multi-graph matching techniques in this context.
In the present study, we extend these preliminary studies in several directions.

First, we introduce an improved simulation framework to generate populations of artificial sulcal graphs and demonstrate their biological plausibility through a quantitative comparison with real data.
Secondly, we benchmark a selection of recently published multi-graph matching techniques against the best pairwise technique for this task (identified in from \cite{Buskulic2021}), and report variations in performances that would clearly impact potential real-world applications, e.g in a clinical context. 
Finally, we compare qualitatively and quantitatively the different graph matching techniques, as well as the previously published approaches Auzias et al. and Kaltenmark et al, on a real data-set of 137 subjects.
In addition, our experiments demonstrate the feasibility of comparing a large population of sulcal graphs based on multi-graph matching techniques, in fully acceptable computing times.
All the source code and data will be shared openly upon publication at \url{https://www.github.com/gauzias/sulcal_graphs_matching}.

\section{Formal problem and state of the art}

In this section, we define formally the problem of matching sulcal graphs, as well as the multi-graph framework.
We then give an overview of the different methods proposed in the literature and provide a more detailed description of the multi-graph matching methods included in our experiments.

\subsection{Undirected attributed Sulcal graphs}

\begin{figure}[htb]
    \centering
    \includegraphics[width=0.9\linewidth]{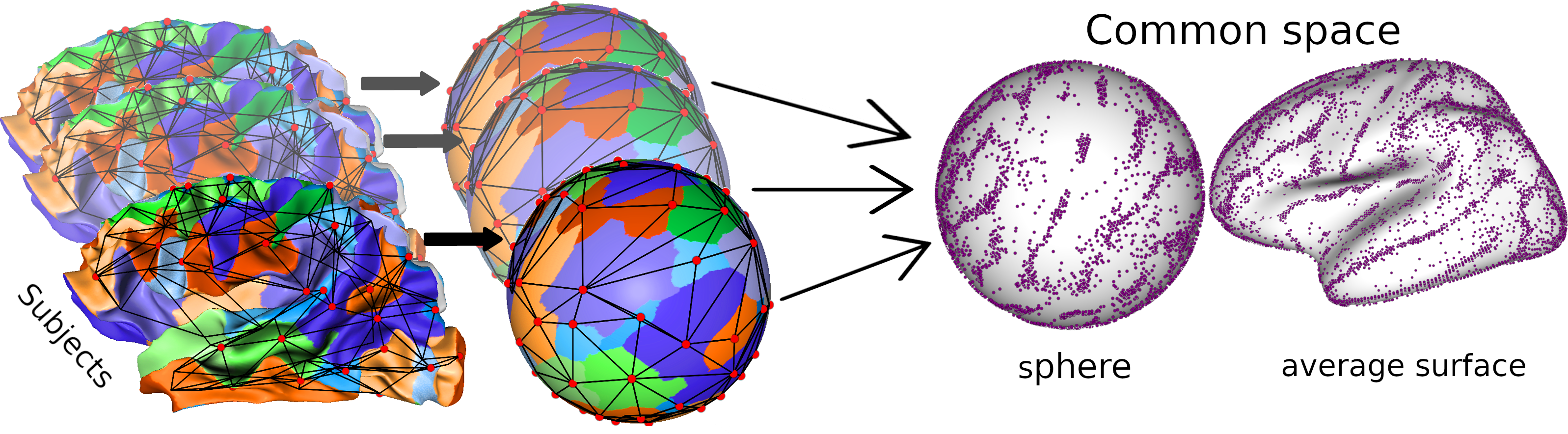}
    \caption{The sulcal graph from each subject is transferred onto a common sphere using the inflation and spherical registration tools from freesurfer. The sulcal graphs from every subjects can then be mapped onto either the common sphere or onto an average surface for visualization. Note that the spatial dispersion of the nodes of the graphs on the common spaces is heterogeneous, with dense clusters in cortical regions where the variations across individuals are lower.}
    \label{fig:subj_to_common}
\end{figure}

We consider a population of $N$ sulcal graphs, noted $\graph_1$ \ldots $\graph_N$, representing the cortical folding pattern of an hemisphere from $N$ different individuals.
The sulcal graph from a given subject $q$ is an undirected attributed graph formally defined as a quadruplet $\graph_q = (V_q, E_q, A^V_q, A^E_q)$, where $V_q = \{v_{1}, v_{2},\ldots,v_{n_q}\}$ are the nodes in the graph and $|\graph_q| = n_q$ is the number of nodes. $E_q \subseteq V_q \times V_q$  defines the set of $e_q$ edges. $A^V_q = \{a^V_{v_1},a^V_{v_2},\ldots,a^V_{v_{n_q}}\}$ is the set of attributes associated to each node in $V_q$, and $A^E_q = \{a^E_{e_1},a^E_{e_2},\ldots,a^E_{e_{e_q}}\}$ is the set of attributes associated with each edge in $E_q$.
Note that the number of nodes $n_q$ and edges $e_q$ and corresponding attributes varies across graphs.
As illustrated on Fig.\ref{fig:subj_to_common}, the sulcal graph from each subject is then mapped onto the same common spherical domain using the surface inflation and registration tools from freesurfer v.5.1.0 (\url{https://surfer.nmr.mgh.harvard.edu/}, see \cite{fischl1999high} for details).
The matching is computed in this common spherical domain. 
In this work, we consider as attributes of the nodes the 3D coordinates of the sulcal pits on the sphere. Regarding the attributes of the edges, we compute the length of the edge on the sphere as an approximation of the geodesic distance between neighboring pits.


\subsection{Generalities and overview of pairwise graph matching methods}
\label{sec:pairwise}

Pairwise graph matching refers to the problem of finding correspondences between the nodes of two graphs $\graph_{1}$ and $\graph_{2}$.
This problem is usually divided into two categories: exact and partial matching. Exact matching methods consider graph matching to be a special case of the graph isomorphism problem. It aims at finding the bijection between two graphs, which implies that both the nodes and edges of the different graphs are strictly matched. This requirement is too strict for most real-world tasks and in particular in our context where the number of nodes and edges varies across graphs. 
Therefore, we focus on the partial matching problem. 
This problem can be formulated as a Quadratic Assignment Problem (QAP) \citep{loiola2007phlebotomine}. Although different forms of QAP exist, the vast majority of the literature has focused on Lawler's QAP~\citep{lawler1963quadratic}. Given two graphs $\graph_{1}$ and $\graph_{2}$ with number of nodes $|\graph_1|=n_{1}$ and $|\graph_2|=n_{2}$ respectively, the Lawler's QAP consists in searching for the \textit{assignment matrix} $\Xmat_{12} \in \{0,1\}^{n_1 \times n_2}$ such that $\Xmat_{12}[i,j] = 1$ indicates that $\upsilon_{i} \in V_{1}$ corresponds to $\upsilon_{j} \in V_{2}$ and $\Xmat_{12}[i,j] = 0$ otherwise, resulting from the following optimization problem:

\begin{gather}
\label{eq:qap_lawler}
   \max{J(\Xmat_{12})} = vec(\Xmat_{12})^\top \mathbf{\Phi}_{12}\, vec(\Xmat_{12})~,\\
    \text{subject to}\ \Xmat_{12} \ones_{n_2} = \ones_{n_1}, \Xmat_{12}^\top \ones_{n_1} \leq \ones_{n_2} , \Xmat_{12} \in \{0,1\}^{n_1 \times n_2}~, \nonumber
\end{gather}
where $vec(\Xmat_{12})$ denotes the column wise vectorization of $\Xmat_{12}$; $\ones_{n_1}$ and $\ones_{n_2}$ denote the column vectors of all ones of size $n_1$ and $n_2$; and $\mathbf{\Phi}_{12} \in [0,1]^{n_1n_2 \times n_1n_2}$ is the \textit{affinity matrix} that is given as an input. 
The diagonal entries of $\mathbf{\Phi}_{12}$ encode the similarity across nodes whereas non-diagonal entries encode the similarity across edges between the two graphs. 
The computation of the affinity matrix is context-dependent, and we detail the approach used in the present work in section \ref{sec:affinity}.

The computation and storage in memory of the very large matrix $\mathbf{\Phi}_{12}$ impedes the scalability of the matching problem based on this formulation.
A solution to tackle this limitation is to reformulate the matching as a Koopmans-Beckmann's problem \citep{zhou2015factorized} that is a special case of Lawler's QAP:
\begin{gather}
\label{eq:qap_koopman}
    \max{J(\Xmat_{12})} = tr(\mathbf{\Psi}_{12}^{\top} \Xmat_{12}) + tr(\Amat_{1}\Xmat_{12}\Amat_{2}\Xmat_{12}^\top)~,\\
    \text{subject to}\ \Xmat_{12} \ones_{n_2} = \ones_{n_1}, \Xmat_{12}^\top \ones_{n_1} \leq \ones_{n_2} , \Xmat_{12} \in \{0,1\}^{n_1 \times n_2}~, \nonumber
\end{gather}
where $\mathbf{\Psi}_{12} \in [0,1]^{n_{1} \times n_{2}}$ denotes the affinity matrix \textit{across nodes}, and $\Amat_{1} \in \mathbb{R}^{n_{1} \times n_{1}}$ and $\Amat_{2} \in \mathbb{R}^{n_{2} \times n_{2}}$ are the weighted adjacency matrices of two graphs respectively such that $\Amat[i, j] = w_{ij}$ if edge $(v_{i}, v_{j})$ exists with weight $w_{ij}$ and $\Amat[i, j] =0$ otherwise.
Koopmans-Beckmann's formulation is a special case of Lawler's where the edges can only be weighted by a scalar value (i.e. cannot support a vector of attributes on edges). 
Under this constraint, we can decompose the large matrix $\mathbf{\Phi}_{12}$ into three smaller matrices $\mathbf{\Psi}_{12}, \Amat_{1}$ and $\Amat_{2}$, which provides better scalability than Lawler's QAP.


These two formulations are combinatorial QAPs and are known to be NP-hard problems.
Most methods therefore relax the hard constraints given in Eq.(\ref{eq:qap_lawler}) and (\ref{eq:qap_koopman}) and provide approximate solutions.
Various approaches have been proposed to relax these problems, leading to a variety of graph matching methods.
Discussing these methods is beyond the scope of this work but we refer interested readers to the review \cite{yan2016short}. 

Going back to our specific context, we reported in \cite{Buskulic2021} a benchmark of the pairwise methods SMAC (Spectral Matching with Affine Constraints) \citep{cour_balanced_2007}, IPFP (Integer Projected Fixed Point algorithm) \citep{leordeanu_integer_2009}, RRWM (Reweighted Random Walks for graph Matching) \citep{hutchison_reweighted_2010}, and KerGM (Kernelized Graph Matching) \citep{zhang_kergm_2019}.
We observed that \textbf{KerGM} clearly outperforms the others in our context.
Conceptually, \textbf{KerGM} well suits sulcal graphs as it relies on Frank-Wolfe optimization that allows to follow an optimisation path that respects the constraint on each step. This induces a robustness to the presence of noise in graphs that is crucial in our context.
In the present work, \textbf{KerGM} is included in our benchmark as a representative of pairwise approaches. It is also used to define the initialization of all the multi-graph methods that are introduced in next section.

\subsection{The multi-graph matching problem}
\label{sec:multi_graph_desc}

We now focus on the problem of jointly matching a population of $N$ graphs $\{\graph_{1}, \ldots, \graph_{N}\}$, starting from pairwise assignment matrices $\Xmat_{ij}$ between graphs $\graph_{i}$ and $\graph_{j}$ (computed with \textbf{KerGM} in this work).
The key concept behind multi-graph matching is the \textit{cycle consistency}. This concept states that a matching between two graphs $\graph_{i}$ and $\graph_{j}$ should be the same if we go through an intermediate graph $\graph_{k}$ to create a new mapping. Formally, a perfectly consistent, bijective mapping (every node is matched to one and only one other node) would satisfy :
\begin{align}
\label{eq:consistency}
    \Xmat_{ik} = \Xmat_{ij} \Xmat_{jk}~,
\end{align}
for any $i$, $j$ and $k$ with $i\neq j \neq k$. 
A common way to estimate consistency at the population level is to compute the full bulk assignment matrix $\mathbb{X} \in \{0,1\}^{m \times m}$ with $m = \sum_{q=1}^N|\graph_q|$, that is obtained by assembling all individual pairwise matrices:
\[
  \mathbb{X} =
  \left[ {\begin{array}{cccc}
    \Xmat_{11} & \Xmat_{12} & \cdots & \Xmat_{1N}\\
    \Xmat_{21} & \Xmat_{22} & \cdots & \Xmat_{2N}\\
    \vdots & \vdots & \ddots & \vdots\\
    \Xmat_{N1} & \Xmat_{N2} & \cdots & \Xmat_{NN}\\
  \end{array} } \right]
\]
Intuitively, enforcing the consistency constraint will induce a reduction of the rank of this bulk matrix.
Multi-graph matching techniques can be divided into three categories as follows.

The first category of approaches explicitly aim at minimizing the rank of the bulk matrix using various approaches \citep{pachauri_solving_2013, chen2014near, wang2018multi,hu2018distributable}. For instance, \citep{bernard2019hippi} solves a global optimization problem by using a projected power iterative method, and we detailed further \citep{zhou2015multi}.

The second category of techniques does not explicitly minimize the rank of the bulk matrix but rely on other types of formalization aiming at increasing the consistency across all graphs
\citep{yan_multi_2016,yan2013joint,yan2014graduated,yan_consistency-driven_2015,yan_multi_2016}. 

Finally, the third category corresponds to deep learning approaches that show promising performances in supervised tasks compared to previous methods, but are not suited for unsupervised tasks \citep{zanfir2018deep,wang2020combinatorial,
wang2019learning,wang2021neural,pmlr-v139-yu21d,yu2019learning,wang2020graduated,rolinek2020deep}. 

Some other interesting methods exploit the concept of consistency in order to solve the problem of jointly matching multiple images \citep{rubinstein2013unsupervised, faktor2013clustering, tron_fast_2017, zhou2015flowweb}. However, these do not take into account the connectivity of the graphs.


\subsection{Selection of the methods included in our benchmark}

We used the following criteria to select the methods included in our benchmark:
\textit{(i) Availability of code.} We included only methods for which the authors have made their code openly available in order to avoid reimplementation issues and to ensure the full reproducibility of our results.
\textit{(ii) Methods exploiting graph topology.} We selected the methods that take into account the topology of the graph, which is crucial to exploit the spatial adjacency information encoded in the sulcal graphs.
\textit{(iii) Scalability.} Since we are interested in performing population studies over large sets of individuals, we excluded methods that do not provide acceptable scalability.
\textit{(iv) Unsupervised methods.} Finally, as motivated in the introduction, we focus on unsupervised methods in the present study.

The method that satisfy these selection criteria are \textbf{mALS} \citep{zhou2015multi}, \textbf{mSync} \citep{pachauri_solving_2013} and \textbf{CAO} \citep{yan_multi_2016}. 
We provide a detailed description of each of these methods below.
In our experiments, these multigraph graph-matching techniques will be compared with the pariwise approach \textbf{KerGM}, and with the two methods from the literature specifically designed for labeling sulcal graphs already described in Sec. \ref{sec:abs_GT}: \textbf{Auzias et al.} \citep{auzias_deep_2015} and \textbf{Kaltenmark et al.} \citep{irene_kaltenmark_cortical_2020}.


\subsection{Description of the selected multi-graph matching methods} 
\label{sec:algo_descr}

As described in section \ref{sec:multi_graph_desc}, the general objective of multi-graph matching methods is to match the nodes across several graphs together by enforcing consistency. 



The authors of \textbf{CAO} \citep{yan_multi_2016} propose to maximize the affinity information and impose consistency at the same time instead of considering them separately. They assume that enforcing consistency acts as a regularizer in the affinity objective function, particularly when the matching is ambiguous due to noise. 
The approach is based on the search of an intermediate graph $\graph_{q}$ that allows to optimize the affinity score while progressively inducing consistency.
They introduce the unitary consistency across a set of $N$ pairwise matching solutions $\mathbb{X}$ for a graph $\graph_{q}$ as:
\begin{equation}
\label{eq:cao}
    C_u(\graph_q,\Xset) = 1 - \frac{\sum^{N-1}_{i=1} \sum^{N}_{j=i+1} \norm{\Xmat_{ij} - \Xmat_{iq} \Xmat_{qj}}_F / 2}{n_{q}N(N-1)/2}~,
\end{equation}
where $\norm{.}_F$ is the Frobenius norm.
The authors propose several approaches to balance between consistency and affinity, leading to different variants of \textbf{CAO}. 
In particular, their best algorithm is able to elicit outlier nodes during the optimization, which is highly relevant in our context. 
However, the use of affinity information along with consistency and outlier elicitation increase the computational complexity of the method to $O(N^4)$.
As a consequence, only the least resource-demanding algorithm ${CAO}^{cst}$ did scale with the memory requirements imposed by the size of our graphs and number of subjects in our populations.
We thus refer to that particular version in the rest of this article.
This version of \textbf{CAO} enforces consistency through Eq.\ref{eq:cao}, but ignores the affinity information.




The approach \textbf{mSync} \citep{pachauri_solving_2013} consists in estimating a mapping of each $\Xmat_{ij}$ to a common \textit{universe} of assignment matrices, of size $d$:
\begin{gather}
\label{eq:mSync}
    \max_{\{U_{i},U_{j}\}\in\pazocal{P}} \sum_{i=1}^{N}\sum_{j=1}^{N} \langle U_{i}U_{j},\Xmat_{ij}\rangle~,\\
\text{with }\pazocal{P} = \{ U \in \{0, 1\}^{n_q\times d} \mid U\mathbf{1}_d =  \mathbf{1}_{n_q}\}.
\end{gather}

Since solving eq.\ref{eq:mSync} is intractable in most applications, the authors relax the problem into a generalized Rayleigh problem. They further propose to use a \textit{reference} graph in order to estimate the mapping to the \textit{universe}. In the implementation provided by the authors, the first graph in the collection $\graph_{1}$ is selected as the \textit{reference} graph. 

In \textbf{mALS} \citet{zhou2015multi}, the authors formalize the multi-graph matching as the following low rank matrix recovery problem:
\begin{equation}
\label{eq:mALS}
    \begin{aligned}
    f(\mathbb{X}) &= -\sum_{i=1}^{N}\sum_{j=1}^{N}\langle \mathbf{\Psi}_{ij},\Xmat_{ij} \rangle + \alpha \langle \textbf{1},\mathbb{X} \rangle + \lambda \lVert \mathbb{X} \rVert_{*}~, \\
    & = -\langle \mathbb{K} - \alpha\textbf{1}, \mathbb{X} \rangle + \lambda \lVert \mathbb{X} \rVert_{*}~,
    \end{aligned}
\end{equation}

where, $\langle .,. \rangle$ is the inner product, $\alpha$ controls the weight on sparsity, and  $\mathbb{K} =  \{\mathbf{\Psi}_{ij}\}_{i,j=1}^{N}$ is the set of affinity matrices given as input.
The cycle consistency is induced by the nuclear norm $\lVert \mathbb{X} \rVert_{*}$ that controls for the rank of $\mathbb{X}$ while $\langle \textbf{1},\mathbb{X} \rangle$ favors bijective matchings across graphs.
Importantly, $\mathbb{X}$ is treated as a real matrix such that $\mathbb{X} \in [0,1]$ The matrix is binarized at the end of the optimization process using a threshold value $t$ that is set by default as to $t=0.5$. 
In, addition, the authors leverage the work by \cite{hastie2015matrix} and \cite{cabral2013unifying} for decomposing $\mathbb{X}$ which allows to solve the problem in a lower dimension space using the ADMM method \citep{eckstein1992douglas}.

\section{Generation of a population of synthetic sulcal graphs}
\label{sec:gen_simus}
A primary objective of our work is to investigate and evaluate different multi-graph matching techniques in the context of sulcal graphs. However, as mentioned in the introduction, there is no ground truth matching available for such graphs. 
We tackle this problem by designing a procedure allowing to generate a population of artificial sulcal graphs with correspondences defined by construction.
Such populations of artificial graphs will constitute a ground truth against which the different matching methods can then be benchmarked. 
Generating artificial sulcal graphs for the purpose of a benchmark study induces the two following constraints: 
1) The artificial graphs should be biologically plausible, i.e. they should respect as much as possible the intrinsic properties of a population of real sulcal graphs.
2) The generation of the artificial graphs should be as simple and straightforward as possible in order to facilitate the comparison of the performances obtained in the benchmark study and the interpretation of the differences, i.e. the generation procedure should rely only on a limited number of parameters, and potential biases should be avoided. 
As detailed below, these two contradictory constraints are balanced in the design of our generation procedure.

The procedure is summarized in Algo. \ref{alg:gen_art} and consists in two main steps. First, we generate a set of points on the common spherical domain, that will serve as \textit{reference nodes}. Then, we impose several types of perturbations to this set of reference nodes in order to generate a corresponding population of artificial sulcal graphs, while preserving the correspondences across graphs, i.e. the ground-truth matching.
Such procedure provides the ground truth matching across the population, while controlling for the nature and amount of variations across artificial sulcal graphs (corresponding to different subjects in real data).

\begin{algorithm}
\caption{Procedure to generate a population of artificial sulcal grahs}\label{alg:gen_art}
\begin{algorithmic}
\Require $N, n_{ref}, \kappa, \mu_{pert}, \sigma_{pert}, p $
\State \textbf{Step1: create reference nodes}  \Comment{See Sec.\ref{sec:gen_ref}}
    \For{$j=1..10000$}
        \State Sample $n_{ref}$ points on the sphere
        \State Compute the minimum geodesic distance
    \EndFor
\State Choose the set of points with the largest min distance.
\State \textbf{Step 2: generate a population of sulcal graphs}  \Comment{See Sec.\ref{sec:gen_inds}}
\For{$i=1..N$}
    \State Perturb location of the reference nodes \Comment{See Sec.\ref{sec:pertub_of_nodes}}
    \State Add outliers and suppress some nodes \Comment{See Sec.\ref{sec:gen_outliers}}
    \State Compute the edges of the graph \Comment{See Sec.\ref{sec:gen_edges}}
\EndFor
\end{algorithmic}
\end{algorithm}

\subsection{Generation of a set of reference nodes}
\label{sec:gen_ref}
The first step consists in generating a set of reference nodes on the spherical domain while controlling for two specific distinct parameters : the \textit{number of nodes} noted  $n_{ref}$, that is typically set to match the average number of nodes across a real population, and the \textit{minimum distance between the nodes}.
Indeed, the nodes of the real sulcal graphs cannot be closer to each other than a minimum distance since they correspond to depth maxima that are not located in the immediate proximity of the boundary of sulcal basins (see \cite{auzias_deep_2015} for further description of the extraction of sulcal pits and basins).
As a consequence the spatial distribution of the nodes on the sphere cannot be fully random.
In order to generate this set of $n_{ref}$ points on a sphere with pseudo-random spatial distribution, we adopted a simple brute force approach: we sample a set of $n_{ref}$ points over the surface of the sphere $10000$ times; and we select the set that has the largest minimum geodesic distance between neighbouring points.
As we will show in sec.\ref{sec:simu_params}, $10000$ times is sufficient to get a set of reference nodes with a minimum distance between points that is realistic.
Technically, the uniform sampling of points on the sphere is achieved by generating random rotations of the unit vector as described in \cite{blaser_random_nodate,lefevre_spanol_2018}.
 
At this stage, we have defined on purpose a set of \textit{reference nodes} that matches a real population in terms of size and of minimal distance between nodes.
The next step consists in perturbing the reference nodes in order to generate the population of synthetic sulcal graphs.

\subsection{Generation of an individual sulcal graphs}
\label{sec:gen_inds}

We now add perturbations of different natures to this set of reference nodes in order to obtain a population of artificial sulcal graphs, that corresponds to different subjects. 
These perturbations aim at mimicking the inter-individual variations that are observed in a healthy population, by affecting the features of the nodes and edges, but also the topology of the graphs.
In order to generate a population of $N$ artificial sulcal graphs, these operations are repeated $N$ times independently. 

\subsubsection{Perturbation of the location of the reference nodes}
\label{sec:pertub_of_nodes}

The first step consists in adding random noise to the coordinates of the reference nodes on the sphere, in order to model the inter-individual variability that exists in the location of the sulcal pits. 
We used the von Mises-Fisher ($vMF$) distribution that is an approximation of Gaussian distribution on a sphere \citep{von1964mathematical}. The two parameters  of the  $vMF$ distribution $\mu$ and $\kappa$ can be seen as the equivalent of the mean and of the inverse of the standard deviation ($\kappa \propto 1/\sigma$) for a Gaussian distribution. 
Therefore, we iterate across the reference nodes, and for each reference node, we produce a noisy one by sampling from the distribution $vMF(\mu, \kappa)$, where $\mu$ is the coordinates of this reference node.
We control for the amount of noise on the coordinates of the perturbed nodes through the value of the parameter $\kappa$, that is common to all nodes from the reference set. 
Smaller values for $\kappa$ will induce larger variations across the artificial sulcal graphs within the population. 
Importantly, note that since we perturb each node of the reference set independently, we keep the correspondence between each noisy node and its reference node, which will allow defining our ground truth matching at the population level.

\subsubsection{Addition of outliers and suppression of nodes}
\label{sec:gen_outliers}

Next, we simulate the inter-individual variations in the number of nodes across the sulcal graphs, which is of crucial importance for generating biologically plausible artificial populations. 
The aim is to model both false positive and false negative matchings, i.e. respectively nodes that are present in the reference set but not in a given graph, and nodes that are present in the graph but not in the reference set.
This is achieved by randomly adding a certain number $n_{o}$ of nodes on top of the perturbed nodes -- hereafter called \textit{outlier nodes}, and by deleting $n_{s}$ nodes amongst the perturbed nodes -- hereafter called \textit{suppressed nodes}.
In order to randomly draw $n_{o}$ and $n_{s}$,
we use the $\beta$-binomial distribution $B(\nu, \alpha, \beta)$, which is a distribution of non-negative integers.
The parameter $\nu$ denotes the size of the support of the distribution, i.e the maximal value that can be sampled. The parameters $\alpha$ and $\beta$ can be set so that $B(\nu, \alpha, \beta)$ approximates a Gaussian distribution. We describe the setting of these parameters and precise their link with $\mu$ and $\sigma$ of a Gaussian in Appendix \ref{annex1}.
Since we want the average number of nodes across the population of perturbed graphs $\mu_{simu} $ to match the number of nodes in the reference set $n_{ref}$,
we set $\mu_{o} = \mu_{s} = \mu_{pert} $  and  also $\sigma_{o} = \sigma_{s} = \sigma_{pert}$.
This formulation allows us to control the standard deviation of the number of nodes across the population of artificial graphs with the two parameters $\mu_{pert}$ and $\sigma_{pert}$.

\subsubsection{Construction of the edges}
\label{sec:gen_edges}
The last step consists in constructing each artificial sulcal graph with the sets of perturbed nodes as follows.
We first compute the three-dimensional convex hull of each set of perturbed nodes located on the sphere. This yields a triangulation where only neighboring nodes on the sphere are connected, which is a simple way to simulate the region adjacency graph that is constructed from the sulcal basins in the real data.
However, the average node degree in such triangulations is higher than for real sulcal graphs. Therefore, we finally delete a small percentage $p$ of the edges in these triangulations, in order to obtain artificial graphs which match the average degree of real sulcal graphs.

Note that since the construction of the edges occurs after the previous perturbation steps (perturbations of the location, addition of outlier nodes and suppression of nodes), the resulting artificial sulcal graphs can show variations in their topology across individuals of a population, as we observe in real data, making them biologically-plausible in that respect.


\section{Experiments and results}


\subsection{Computation of the affinity matrices}
\label{sec:affinity}

As described in Sec.\ref{sec:pairwise}, we initialize all the multigraph matching methods using the pairwise results obtain from \textbf{KerGM}, which relies on the formalization of Eq. \ref{eq:qap_koopman}.
We thus need to compute the affinity matrices $\mathbf{\Psi}_{ij}, \Amat_{i}, \Amat_{j}$ that store the similarity between nodes and edges across every pairs of graphs in the population. 

In the present work, we compute these affinity matrices using Gaussian kernels applied to the attributes. For two nodes $v \in \graph$ and $v' \in \graph'$ the affinity value is computed using the kernel defined as $\exp{(-\gamma^V\norm{a^V_v - a^V_{v'}}_2^2)}$ and for two edges $e \in \graph$ and $e' \in \graph'$ the kernel is defined as $\exp{(-\gamma^E\norm{a^E_e - a^E_{e'}}_2^2)}$. To estimate appropriate values for $\gamma^V$ and $\gamma^E$ we use a heuristic proposed in \cite{takerkart_structural_2017} that consists in using a cross-validation scheme to compute the inverse of the median of the distribution across all possible pairs of nodes/edges, independently for each attribute (3D coordinates on the sphere for the nodes and the geodesic distance for the edges).


\subsection{Dummy nodes}
Most graph matching methods assume a constant number of nodes across the graphs to be matched, which is not the case in our case (both synthetic and real graphs).
We use the classical approach from the graph matching literature which consists in adding \textit{dummy nodes} to smaller graphs so that all the graphs get the same number of nodes as the largest graph in the population. 
For each of these dummy nodes, we assign to 0 the corresponding values in the node and edge affinity matrices. 
This makes the optimization problem defined in Eq.\ref{eq:qap_koopman} independent from dummy nodes.

\subsection{Benchmark on synthetic sulcal graphs}
\label{sec:res_bench}
\subsubsection{Description of synthetic data sets}
\label{sec:simu_params}

We first tuned empirically the parameters to the values $ \mu_{pert} = 12, \sigma_{pert} = 4$  and $p=10\%$ to obtain variations in our synthetic graph populations that are in line with what is observed in real data. The distribution for number of nodes in the real data population is $88.27 \pm 4.72$ likewise in our simulated population for a randomly chosen $\kappa$ value the distribution for number of nodes is $88.15 \pm 4.45$ for the selected value of $\mu_{pert}$ and $\sigma_{pert}$ and is consistent across all $\kappa$ values across all trials. We further provide in Appendix \ref{annex2} additional materials showing the matching distributions between our simulated graphs and real data.

Furthermore, we varied the value of $\kappa \in [100,200,400,1000]$, which controls the amount of variations across synthetic graphs within a population.
Note that $\kappa$ controls the spread of nodes coordinates around the reference nodes, which in turn induces variations in the topology and attributes of synthetic graphs.

\begin{figure}[!h]
    \centering
    \includegraphics[width=1.0\linewidth]{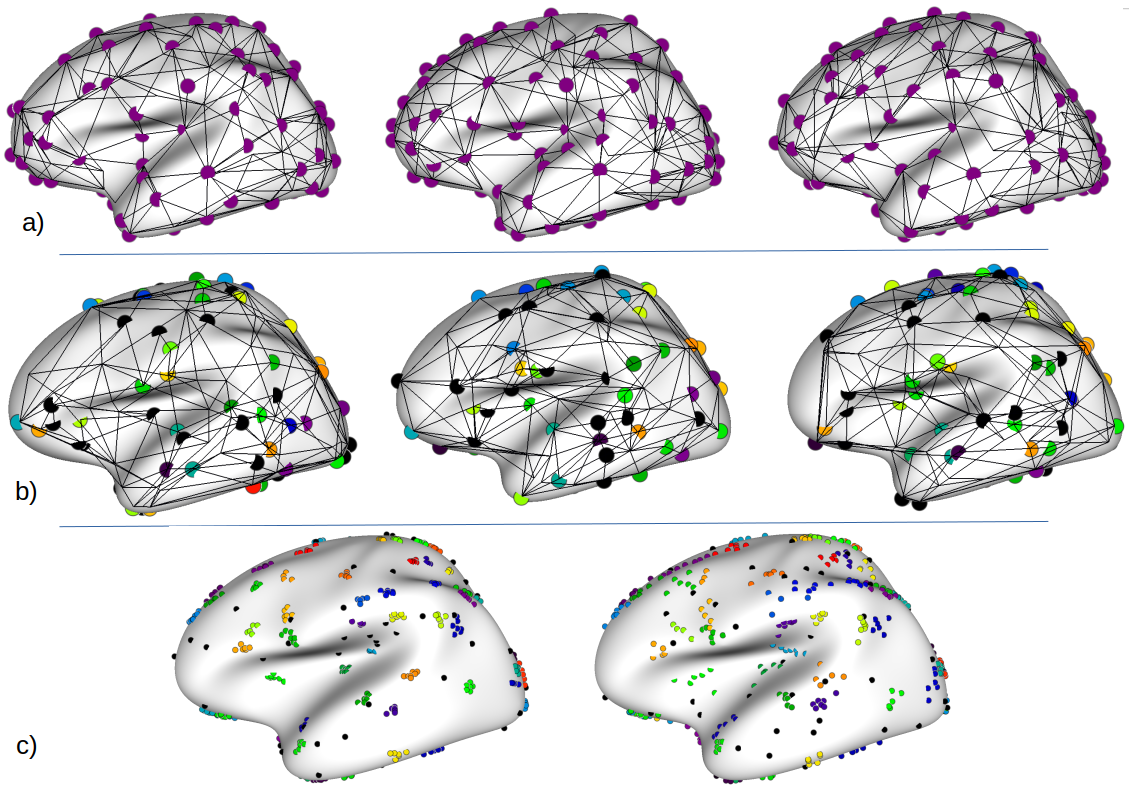}
    \caption{a) Real sulcal graphs from three randomly chosen individuals, and projected on the average surface.
    b) Simulated graphs randomly chosen for $\kappa = 1000$, showing the ground-truth correspondence across graphs in color. Nodes in black represent the outlier nodes that have no correspondence.
    c) Illustration of the impact of $\kappa$ on the spatial dispersion of nodes: the nodes of six simulated graphs are shown on the average surface for $\kappa=1000$ (left) and , $\kappa=200$ (right). The spread across the nodes for each cluster varies according to $\kappa$, while outlier nodes in black have random locations.}
    \label{fig:visu_simu}
\end{figure}

For each value of $\kappa$, we generate 10 populations of $N=137$ synthetic graphs (which corresponds to the number of subjects in our real population; see below) and report the average and standard deviation of the metrics described below.
As illustrated on Fig.\ref{fig:visu_simu}, our populations of synthetic graphs show variations that are qualitatively very close to those observed across real graphs.

\subsubsection{Evaluation metrics for synthetic data sets}

In order to evaluate the different matching methods on simulated graphs, we use the classical \textit{precision}, \textit{recall} and $F_1$\textit{-score}:

\begin{align}
    \textit{Precision} =\frac{\text{\textit{True~Positives}}}{\text{\textit{True~Positives} + \textit{False~Positives}}} \in [0,1]
\end{align}

\begin{align}
    \textit{Recall} =\frac{\text{\textit{True~Positives}}}{\text{\textit{True~Positives} + \textit{False~Negatives}}} \in [0,1]
\end{align}

\begin{equation}
   F_1 = 2\frac{(precision \times recall)} {precision+recall}  \in [0,1]
\end{equation}

Thus, \textit{Precision} is a ratio between the True positives(\textit{number of correct matches predicted by the algorithms}) and all the positives\textit{(number of matches by the algorithms)}. Whereas, \textit{Recall} is a ratio between True positives and True positives along with False negatives(\textit{number of correct matches not predicted by the algorithms}). Finally, the $F_1$ score provides a balance between \textit{Precision} and \textit{Recall}. A $F_1$\textit{-score} of $1$ reflects the ability of the algorithm to obtain a perfect matching of inlier nodes and accurate identification of outlier nodes.
These metrics are relevant in our context to detect matching with outliers alongside the incorrect matches.


\subsubsection{Results on synthetic data sets}
\label{sec:res_syn}

\begin{figure}[h!]
    \centering
    \includegraphics[width=1.0\linewidth]{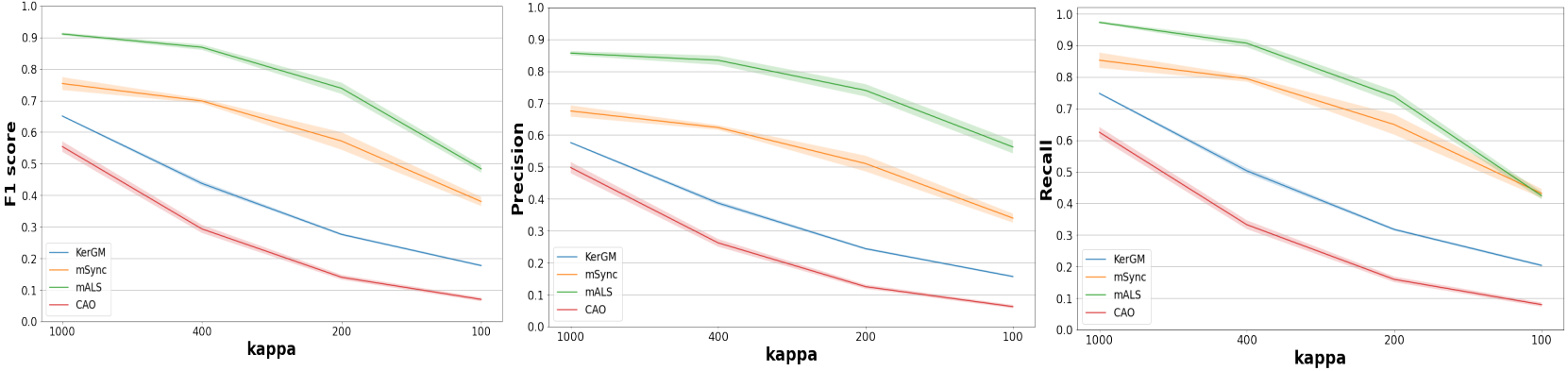}
    \caption{$F_1$-score, Precision and recall for $\kappa \in [1000,400,200,100]$. For each method, we plot the average across the 10 simulated populations as a line and the standard deviation as the shaded region of the same color.}
    \label{fig:benchmark}
\end{figure}

We report of Fig \ref{fig:benchmark} the mean and standard deviation of \textit{Precision}, \textit{Recall} and $F_1$\textit{-score}, computed across the 10 synthetic populations for each value of $\kappa$.

First, we find that two multi-graph matching methods, \textbf{mALS} and \textbf{mSync}, vastly and consistently outperform \textbf{KerGM}, which has been identified as the best pairwise matching method for this task in \cite{Buskulic2021}. This confirms our main hypothesis: considering the matching problem on the whole population using multi-graph matching allows an important gain in performance compared to only considering pairs of graphs.

Then, we observe a gradual decline in the performances of all methods as the noise increases (decrease of $\kappa$), as expected.
The performances of the multigraph approaches \textbf{mALS} and \textbf{mSync} resist much more to this increase in variability than the pairwise approach. 
The performances of \textbf{mSync} are limited more specifically by the lower precision at any level of noise.
This suggests that the difference in performances between the two methods are mainly due to the hard constraint on the consistency in \textbf{mSync} that seems too restrictive. 
On the other hand, the recall indicates that \textbf{mSync} is more robust to increasing noise than \textbf{mALS}, with very close value when $\kappa=100$.
However, \textbf{mALS} performs better for lower noise values.
Overall, \textbf{mALS} shows the best $F_1$\textit{-score} for every $\kappa$ values, thanks to a very high precision combined with very good recall. Indeed, the $F_1$\textit{-score} for \textbf{mALS} is above 0.7 even for $\kappa=200$ which corresponds to a configuration where the noise is quite strong. 

Finally, the performances of \textbf{CAO} are very low, even lower than the pairwise technique \textbf{KerGM}. 
Such poor performances are likely a consequence of the optimization that considers only the consistency but ignores the affinity of nodes.
As already mentioned in Sec.\ref{sec:algo_descr}, the other versions of \textbf{CAO} proposed in \cite{yan_multi_2016} could show much higher performances but did not scale with the size of our data.

\subsection{Application to real data}
\label{sec:real_data}

\subsubsection{Preprocessing of real data}
For the evaluation on real data, we use the sulcal graphs from 137 young healthy adults taken from the publicly available database OASIS \citep{marcus2007open}. 
The preprocessing of these data (brain tissues segmentation, mesh extraction and sulcal graphs construction) has been detailed in \citet{auzias_deep_2015, takerkart_structural_2017}. 
Across this population, the number of nodes is $88 \pm 4$, with a maximum size of $101$ nodes/pits. Dummy nodes are thus added to all other graphs to get a constant size of 101, as explained above.

\subsubsection{Evaluation metrics used with real data}
\label{sec:real_metrics}
In absence of ground truth matching for real data, we cannot compute the same scores as for the simulation experiments.
We therefore combine a set of quantitative metrics with some qualitative assessments, which we describe below.

\textbf{\textit{Consistency}}

According to \cite{yan_multi_2016}, we compute the node consistency as follows:
\textit{Given $\graph_k \in \{\graph_q\}_{q=1}^{N}$ and the bulk matrix $\mathbb{X}$, for node $v^k \in  \graph_k$, with index $i(v^k) \in \{1, \ldots, |\graph_k|\}$, its consistency is defined by:}
\begin{equation}
C(v^k,\mathbb{X})= 1-\frac{\sum_{i=1}^{N-1}\sum_{j=i+1}^{N}||\Ymat(v^k,:)||_F/2}{N(N-1)/2},  \in (0,1], 
\end{equation}
\textit{where $||\cdot||_F$ is the Frobenius norm, $\Ymat=\Xmat_{kj}-\Xmat_{ki}\Xmat_{ij}$ and $\Ymat(v^k,:)$ is the $i(v^k)$-th row of matrix $\Ymat$.}
Note that it is different from Eq. \ref{eq:cao} which estimates the consistency at the graph level. 
This consistency measure is computed for each node of each graph, including dummy nodes.
A value of 1 corresponds to the ideal case where each graph only contains nodes that have been matched in a consistent manner.
This consistency measure cannot distinguish the matches of real nodes to dummy nodes from valid matches across real nodes.
For methods imposing an explicit constraint on the consistency, a value of 1 is expected (and not informative), but for the other methods this measure is relevant and allows to assess the spatial pattern of the consistency across clusters.

\textbf{\textit{Qualitative and quantitative assessment of the labeling induced by the matching}}

In terms of potential applications of the graph matching to sulcal graphs, a major outcome is the labeling of graph nodes that is induced.
As already mentioned in the introduction, the assessment of the quality of the labeling and thus of the biological relevance of the matching across individuals is an ill-posed problem.
The first problem is to retrieve a labeling from the assignment matrix resulting from the matching.
In the case of a perfectly consistent matching where each node of each graph would be matched to one and only one node from every other graph in the population, the labeling would be trivial and would consist in simply associating a label to each row or column of the assignment matrix.
This situation is however impossible since the number of nodes varies across individuals within our population of interest.
Therefore, in the present work we take the largest graph as a reference, and we associate an arbitrary label to each of its nodes and then propagate these labels to every other graphs based on the assignment matrix resulting from each method.   

Once the labeling of the nodes is retrieved, the nodes that share the same label across subjects are grouped together into what we will designate as \textit{clusters}, that are different depending on the matching method.
We then compute the coordinates of the centroid of each cluster, which enables to evaluate qualitatively the spatial distribution of the different clusters across the cortical surface.

This qualitative assessment is complemented with a quantitative measure of the compactness of the clusters.
For this, we compute the silhouette coefficient of each node from each graph.
As proposed in \cite{rousseeuw_silhouettes_1987}, the silhouette of a node corresponds to the ratio between the average Euclidean distance to the other nodes in the cluster and its distance to other nearby clusters. Since the distances are computed on the spherical domain, the use of Euclidean distance is sub-optimal but the errors induced are very low and independent from the matching method.
The silhouette coefficient of a cluster is then obtained by averaging the silhouette values from corresponding nodes.


\subsubsection{Results on real data}
\label{sec:res_real}
We first report in Table~\ref{table1} the quantitative measures that allow us to compare the different techniques at the whole brain level: the number of clusters (thus of labels) obtained with each method, the silhouette measure averaged across all nodes and graphs, the percentage of nodes remaining unlabeled, the consistency measure averaged across all nodes and graphs, and the computing time.

\begin{table}[h!]
    \begin{center}
	\caption{Quantitative measures computed at the whole brain scale.}\label{table1}
	\scalebox{0.8}{%
	\begin{tabular}{c| c| c| c| c| c}
		\hline
		\hline
		Method & Num. & silhouette & Perc.       & consistency & cpu time\\
		       & clusters  &            &  unmatched  &             & (min) \\
		\hline
		\textbf{mALS} & 82 & $0.55\pm0.22$ & $28.4$ & $0.91\pm 0.08$ & 783\\
        \textbf{Kaltenmark et al} & 94 & $0.44\pm0.23$ & $17.0$ & $1.0\pm 0.0$ & $\sim 180$ \\
        \textbf{Auzias et al} & 104 & $0.49\pm0.18$ & $0$ & $0.82\pm0.15$ &  $\sim 30$\\
        \textbf{mSync} & 101 & $0.08\pm0.49$ & $0$ & $1.0\pm0.0$ & 31\\
        \textbf{CAO} & 101 & $-0.12\pm0.45$ & $0$ & $1.0\pm0.0$ & 3255 \\
        \textbf{KerGM} & 101 & $-0.04\pm0.34$ & $0$ & $0.30\pm 0.17$ & 1362\\
		\hline
	\end{tabular}}
\end{center}
\end{table}

The number of clusters and percentage of unmatched nodes indicate that the two methods that allow partial matching \textbf{mALS} and \textbf{Kaltenmark et al.} result in a lower number of clusters, suggesting that the ambiguous nodes remain unlabeled instead of enforcing their matching into potentially unreliable clusters.
The three methods \textbf{mSync}, \textbf{CAO} and \textbf{KerGM} enforce the matching of every nodes, and result in a number of clusters equal to the size of the largest graph in the set, i.e. 101.
The method \textbf{Auzias et al.} results in more clusters than the size of the largest graph, suggesting that some clusters correspond to highly variable nodes that cannot be matched consistently across individuals.
This is confirmed by the consistency measure which is lower than for \textbf{mALS}.
The consistency of \textbf{Auzias et al.} is still much higher than the value of $0.30$ obtained with the pairwise technique \textbf{KerGM}.
Note that the methods \textbf{mSync} and \textbf{CAO} explicitly enforce a perfect consistency, but this is possible only when considering the dummy nodes as pointed in section \ref{sec:real_metrics}. 
Also note that the method \textbf{Kaltenmark et al.} also gets a perfect consistency.
This is a consequence of the explicit constraint imposed in this technique by allowing one and only one node per subject to be matched for any given cluster.

The silhouette measures illustrate that a high consistency can be associated with a low compactness of the clusters as e.g. for \textbf{CAO} and \textbf{mSync} that get values close to the one of the pairwise technique \textbf{KerGM}.
The methods \textbf{Auzias et al.} and \textbf{Kaltenmark et al.} get much higher silhouette values which is expected since these techniques enforce the matching of nodes based essentially on their spatial proximity on the surface.
The silhouette value of \textbf{mALS} is higher than these two techniques.
Overall, \textbf{mALS} results in high silhouette and consistency values, at the cost of a high number of unmatched nodes (28.4\%) compared to \textbf{Kaltenmark et al.} and \textbf{Auzias et al.}, indicating that this method was much more conservative in the matching, leaving more ambiguous nodes unmatched.

\begin{figure}[ht!]
    \centering
\includegraphics[width=\linewidth]{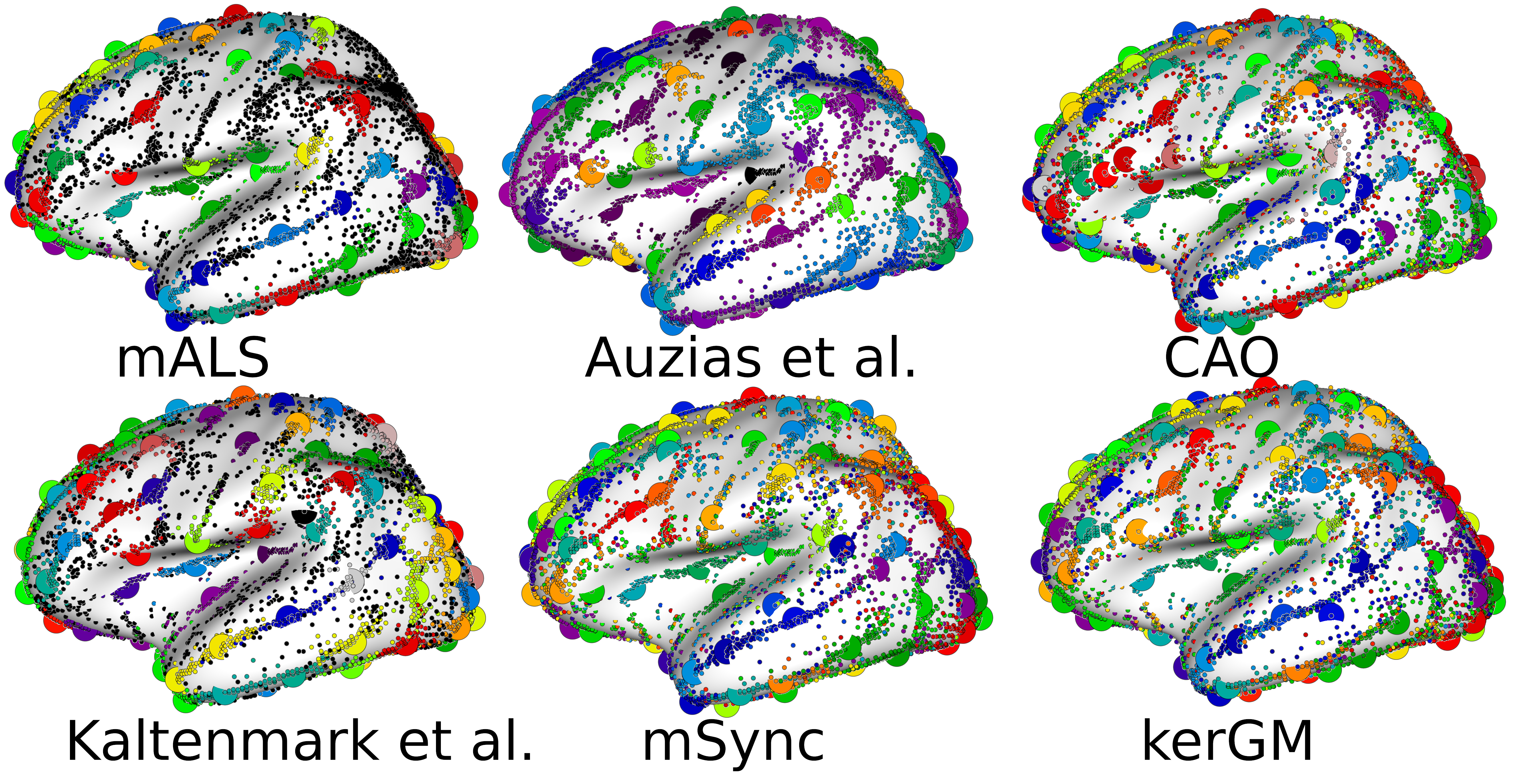}
    \caption{Labeling and corresponding cluster centroids (larger circles) for each method. Dots in black in the first column (for \textbf{mALS} and \textbf{Kaltenmark et al.}) correspond to unmatched nodes. See text for further description.}
    \label{fig:labeling}
\end{figure}

We then illustrate the matching across nodes from the different graphs (subjects), obtained for each method on Fig.~\ref{fig:labeling}.
The number and location of the different centroids (larger circles) is informative of the spatial distribution of the clusters of nodes across the cortical surface, for each method.
On the first column (\textbf{mALS} and \textbf{Kaltenmark et al.}) some nodes remain unlabelled and are represented in black. The clusters seem more compact than for the methods of the second column (\textbf{Auzias et al.} and \textbf{mSync}) that do not allow any node to remain unlabeled. On the third column (\textbf{CAO} and \textbf{kerGM}) the matching looks noisy, with clusters overlapping between eachother in almost every cortical location, which illustrates the poor anatomical relevance of the matching.

\begin{figure}[h!]
    \centering
    \includegraphics[width=\linewidth]{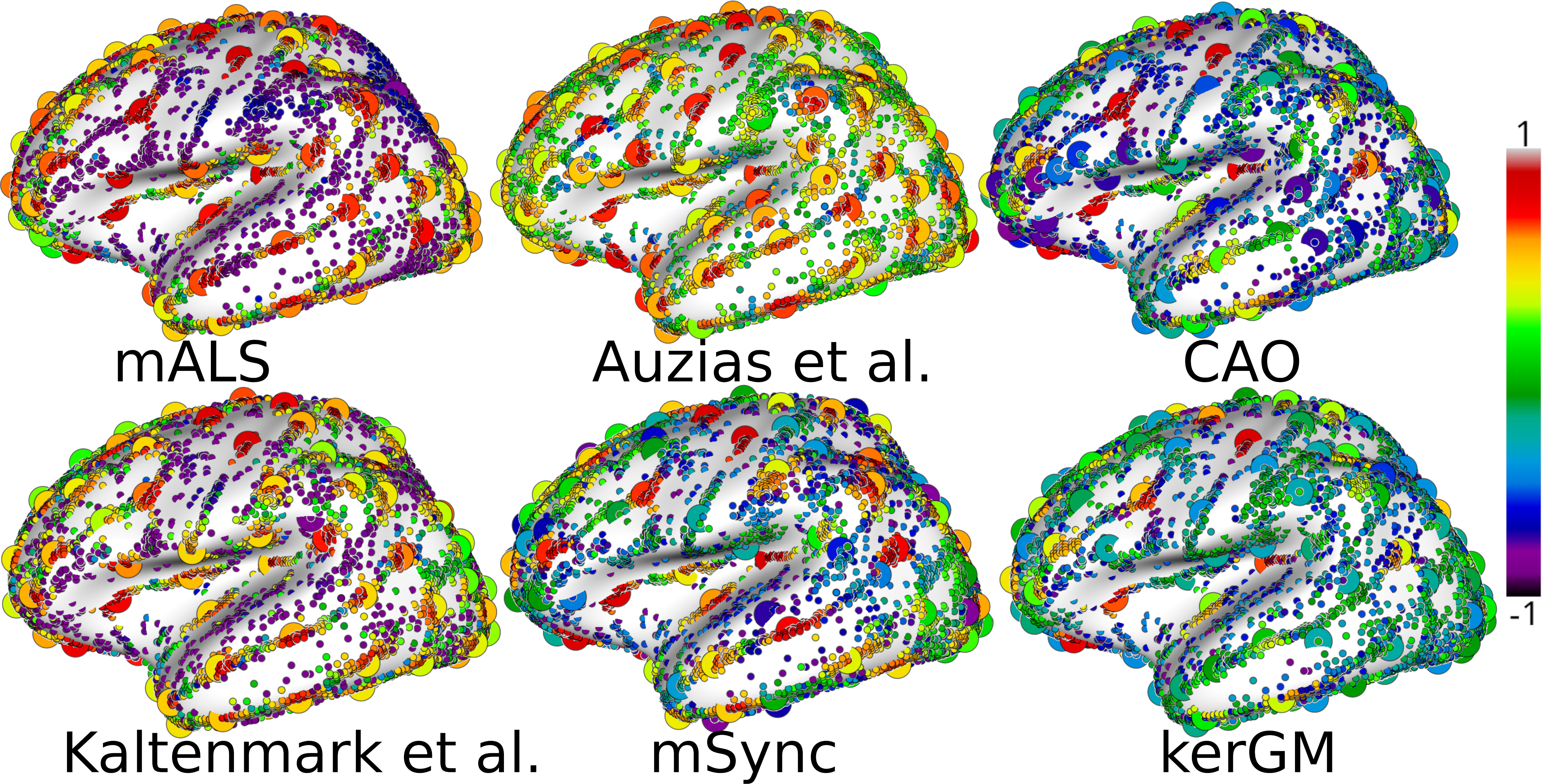}
    \caption{For each method, we show the silhouette coefficient of each node from every graphs, as well as corresponding centroids as larger circles. Each centroid (larger circles) is colored according to the average of the silhouette coefficient of corresponding nodes.}
    \label{fig:silhouette}
\end{figure}

For further evaluation of the performances of the different techniques, we show on Fig.~\ref{fig:silhouette} the silhouette values of every nodes across all graphs as well as the centroids of each cluster as a larger circle. The high silhouette values of the centroids for the methods \textbf{mALS}, \textbf{Auzias et al.} and \textbf{Kaltenmark et al.} are visible with mostly red and orange centroids.
In contrast, we observe more centroids in green and blue for \textbf{CAO} and \textbf{KerGM}.
Together with Table \ref{table1}, this figure illustrates the poor performances of pairwise matching approach with high spatial dispersion of nodes corresponding to each cluster for \textbf{KerGM}, associated to very low silhouette coefficients.
The method \textbf{mSync} results in higher silhouette coefficients for some nodes, but lower value for others (nodes and centroids in blue on Fig.~\ref{fig:silhouette}), indicating that the matching was enforced also for ambiguous nodes located in highly variable regions.
This is a consequence of the hard consistency constraint in \textbf{mSync} imposing a matching that is consistent across all graphs by construction, even in highly variable regions. 
For \textbf{Auzias et al.}, we observe that the clusters are organized around regions of high nodes density, but the nodes located relatively far from the centroids have a lower silhouette value (nodes in green on Fig.~\ref{fig:silhouette}). 
These observations are consistent with the algorithm that is based on a watershed applied to the sulcal pits density map as described in Sec.\ref{sec:abs_GT}.  
For both \textbf{mSync} and \textbf{Auzias et al.}, we observe some clusters with low silhouette value located close to each other, suggesting that the number of clusters is too high.

The techniques \textbf{mALS} and \textbf{Kaltenmark et al.} result in much higher silhouette values, which is expected since they do not force the matching of highly variable nodes that are left unlabeled.
The unlabeled nodes have a very low silhouette value (in violet on Fig.~\ref{fig:silhouette}), but since they do not belong to any cluster, 
this does not reduce the silhouette values of clusters.
Note that even for these methods, the clusters get closer with lower silhouette values in highly variable regions such as the anterior frontal and occipital lobes.

Across the different methods, we observe that the clusters showing a higher silhouette value relative to other clusters are located systematically in the same regions that are known to be less variable across individuals, such as the central sulcus, and the insula. 
For these clusters, the silhouette values are close across methods, confirming the lower ambiguity in the matching in these regions.
In highly variable regions, the different methods produce different matchings.
For instance in the occipital lobe, the clusters produced by  \textbf{Kaltenmark et al.} show lower silhouette values compared to \textbf{mALS}, but we observe the opposite effect in the anterior frontal lobe. 

\begin{figure}[h!]
    \centering
\includegraphics[width=\linewidth]{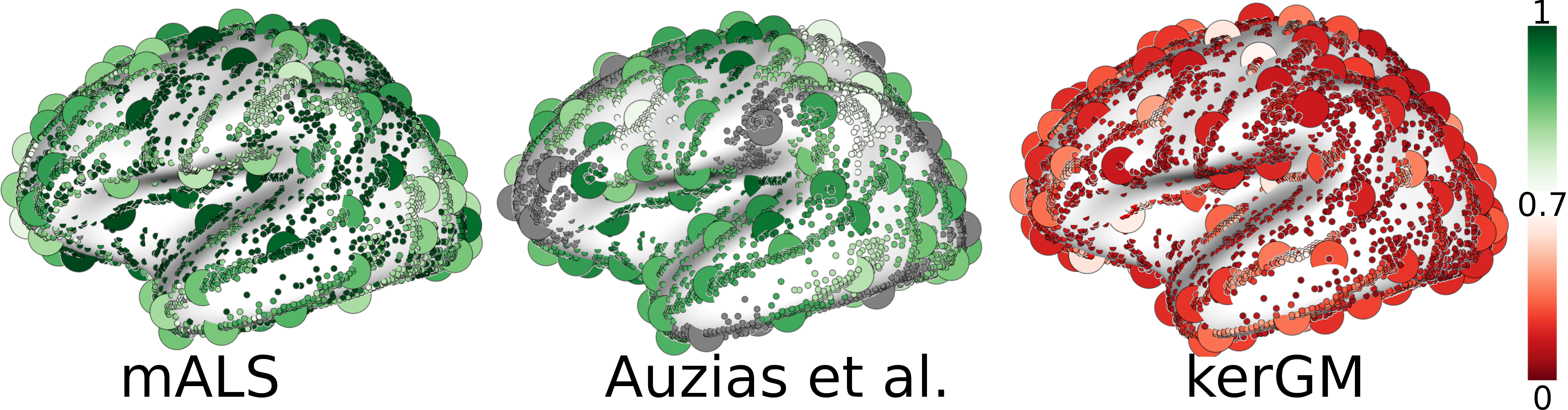}
    \caption{Node consistency computed for each node of each graph with respect to the remaining graphs, and then averaged across graphs. We adapted the colorbar to visualize the differences between the three methods, with the pariwise technique \textbf{KerGM} showing much lower values.}
    \label{fig:consistency}
\end{figure}

On Fig.~\ref{fig:consistency}, we show the consistency for every nodes and centroids, for the three methods that do not explicitly enforce a perfect consistency.
Clearly, the pairwise technique \textbf{KerGM} results in inconsistent matching for every clusters, including the regions where the variations across individuals are known to be low (no centroid in green, even in the central sulcus and the insula).
For \textbf{mALS} and \textbf{Auzias et al.}, we can observe the spatial variations of the consistency across cortical regions.
Again, higher consistency is obtained in less variable regions (central sulcus, insula) for both techniques, and relatively lower values are visible in the frontal and occipital regions.
The consistency is higher for \textbf{mALS} than \textbf{Auzias et al.} for every clusters.
Note that the spatial pattern of the consistency measure for \textbf{mALS} is anatomically relevant, with a consistent matching in the insula, the  central and pre-central regions, and less consistent in the peri-sylvian regions.
At a more local scale, we observe a cluster in the superior temporal sulcus that is more consistent than those located anteriorly or posteriorly, which is in line with previous studies describing variations and stabilities across individuals in this region \cite{leroy_new_2015}.

\section{Discussion}
In this work, we explored the potential of graph matching methods applied to a population of sulcal graphs to uncover correspondences across individuals driven by the local patterns of folds.
Indeed, these graph matching methods take into account the characteristics of individual sulcal basins as well as their topological organization to construct the correspondences.
Our results on both simulations and real data support the biological relevance of the correspondences across individual resulting from multi-graph matching techniques.

\subsection{Relevance of simulated graphs relative to real data for evaluating matching techniques}

To overcome the lack of ground truth for real data, we proposed a procedure allowing to generate artificial graphs that approximate the features of real sulcal graphs while controlling the variations across graphs. 
This simulation procedure enabled to benchmark various pairwise and multi-graph matching techniques. 
The evaluation of the performances of the different methods and their robustness to controlled variations in the simulated graphs was informative for probing their effectiveness in this context.
The performances of the pairwise approach \textbf{KerGM} were limited even when the level of perturbations was minimal.
Note that we reported in \cite{Buskulic2021} that alternative pairwise techniques perform even worse on this task.
Amongst the different multi-graph matching techniques that were tested, \textbf{mALS} showed better performances than the others in all conditions, and a good robustness to increasing noise levels.
These observations were confirmed by our application on real data.
Overall, our set of experiments confirmed the intuition that multi-graph matching techniques are highly relevant in our context, while pairwise techniques show limited performances and might thus be restricted to initialization purpose. 

Of note, our aim was not to push the biological plausibility of our simulated graphs.
Keeping the simulations simple enables straightforward interpretation of the variations in the performances across the different approaches.
This trade-off is visible in the procedure in particular when we sample the reference nodes uniformly on the sphere.
Indeed, our simulation procedure cannot produce realistic non-uniform spatial distribution of nodes across the population. 
While this could be achieved by adapting the sampling of the reference points, this would induce variations in the performances of the matching techniques depending on the location on the sphere, which in turn would make the comparison across methods much more difficult.

Beyond the present work, our procedure for simulating sulcal graphs could be instrumental to assess future improvements in graph matching techniques.

\subsection{Considerations relative to deep-learning approaches and potential methodological improvements}

As already mentioned in section \ref{sec:multi_graph_desc}, many other graph matching techniques can be found in the literature but were not included in the present work.
More specifically, deep learning approaches outperform traditional approaches in supervised learning task \citep{lecun2015deep}. 
Recent works such as e.g. \citep{GNN,xu2018how} showed that the structural information can be learnt by a Graph Neural Network(GNN), providing that manually labelled ground-truth data is available. 

In addition, the rise of semi-supervised learning approaches represents an opportunity in the context of graphs with partial matching ground-truth. Such approaches are worth considering in our context, since we observed marked variations across cortical regions in the ambiguity of the matching. The work by \cite{Fey2020Deep} considers a semi-supervised framework for handling the matching problem where the ground-truth correspondence are only given for a small subset of nodes. In addition, their approach imposes an explicit inductive bias to find correspondences across graphs, based on neighbourhood consensus that does not allow adjacent nodes from being mapped to different regions in other graphs. 
This is appealing in the case of sulcal graph matching where we would like to enforce the matching of nodes located in some specific regions more than in others. 
Such a framework could benefit from the recent work \cite{lyu_labeling_2021} on context-aware data augmentation, which could be instrumental to overcome the bottleneck of the lack of ground-truth labeling data.

Another avenue for potential gains in performance consists in improving the definition and integration of the attributes on nodes and edges.
Many other geometrical features could be considered to enrich the attributes on nodes, such as e.g. shape index and curvedness \cite{awate_cerebral_2010}, or the local gyrification index \cite{rabiei_local_2017}.
On the other hand, the literature on learning edge representations is very scarce \cite{hsu2022graph}, and the attributes on edges are most often reduced to a scalar value (i.e. a simple weight).
In particular, the methods included in the present work cannot handle \textit{vectors} of attributes on edges.
Some recent deep learning methods such as \citep{wang2021neural} can exploit such vectors of attributes, but their scalability is limited by the size of the affinity matrices.
We proposed in \cite{dupe22} to overcome this limitation by leveraging the recent matrix factorization method from \cite{zhang_kergm_2019}. 
We will further investigate the potential of these methods in our future studies.

\subsection{Data-driven nomenclature of sulcal basins}

\begin{figure}[ht!]
    \centering
\includegraphics[width=\linewidth]{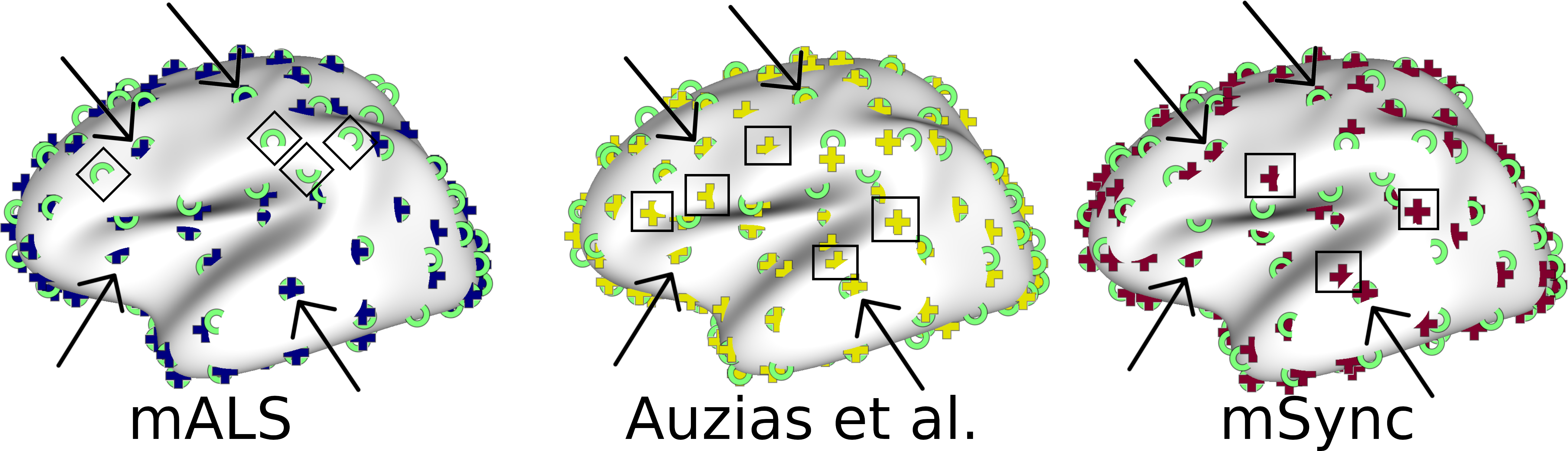}
    \caption{Superimposition of the centroids from \textbf{mALS}, \textbf{Auzias et al.}, and \textbf{mSync} shown as crosses with those of \textbf{Kaltenmark et al.} shown as green rings. \textbf{mSync} is representative of the methods \textbf{kerGM} and \textbf{CAO} that also result in 101 clusters. The arrows point to centroids that are robust across methods. The squares indicate centroids corresponding to small clusters located on gyri. Diamonds indicate centroids that differ between \textbf{mALS} and \textbf{Kaltenmark et al.}.}
    \label{fig:cluster_location}
\end{figure}

The present work extends previously reported experimental results illustrating the major impact of the labeling strategy on the induced correspondences and data-driven nomenclature.
In \cite{irene_kaltenmark_cortical_2020}, the number of clusters obtained at the group level varied from 90 to 114 for the right hemisphere using either the approach proposed in \cite{irene_kaltenmark_cortical_2020} or \cite{auzias_deep_2015} respectively, on the same population of subjects.
We provide a much more detailed comparison.
We show on Fig.~\ref{fig:cluster_location} the superimposition of the centroids from different methods on the same average surface.
This visualization shows that the location of some of the centroids are very consistent across methods (indicated by arrows), corresponding to cortical regions where variations across individuals are known to be low, such as the central sulcus, the insula, the inferior precentral or superior temporal sulcus.
Other clusters differ across methods.
The clusters indicated by squares are those resulting from \textbf{Auzias et al.} and \textbf{mSync} (resp.) that do not match clusters from \textbf{Kaltenmark et al.} (see also Table \ref{table1}).
These are located either in highly variable regions such as the frontal lobe, or on the top of gyri such as the superior temporal gyrus and the inferior frontal gyrus. 
While \textbf{mALS} and \textbf{Kaltenmark et al.} result in centroids that are highly similar (crosses and rings are often superimposed on the panel on the left), the two methods do not result in the same matching in highly variable regions such as the inferior frontal and parietal regions (indicated by diamonds).
In conjunction with our results on synthetic (Sec.\ref{sec:res_syn}) and real data (Sec.\ref{sec:res_real}), these observations confirm that the conceptual differences between the approaches yield different matchings and thus different correspondences across individuals.
Indeed, graph-matching techniques such as \textbf{mALS} are able to take into account the topological information encoded in the graphs, i.e the spatial organization of neighbouring folds, while \textbf{Kaltenmark et al.} relies only on the geometry of sulcal basins, considering different folds separately.


The next step will be to assess the biological relevance of the induced correspondences across subjects by visualizing the matching on the cortical surface of the individuals.
Given the variations across methods in the location of clusters observed on Fig.\ref{fig:cluster_location}, we expect to observe important differences between the techniques at the individual level, especially in highly variable regions such as the parietal lobe.
More specifically, our expectation is that graph matching techniques should allow solving potential anatomical ambiguities in a much more relevant way than \textbf{Auzias et al.} and \textbf{Kaltenmark et al.}, by exploiting the topological information of the neighbouring folding pattern.


\section{Conclusion}

In this study, we explored the potential of several graph matching methods chosen from the literature to define population-wise correspondences across individual cortical geometries. 
In the absence of a ground-truth labeling for real data, we first 
proposed a procedure to generate simulated sulcal graphs that follow the intrinsic structure and properties of real sulcal graph. 
We then compared the approaches on our simulated sulcal graphs with ground-truth correspondences defined by construction. 

We also evaluated the methods on real data. 
We computed the silhouette value of each node of the graph that measures the degree of compactness of each cluster, giving us insights on the matching across graphs produced by the different methods. 
We also computed a consistency measure that gave us an insight on the variability across the population for each cluster. 
The results obtained on real data were compared with two other methods from the literature.

Overall, our experiments on both artificial and real data showed the high relevance of multi-graph methods for sulcal graph matching.
We observed that \textbf{mALS} and \textbf{mSync} outperform \textbf{CAO} and the pairwise approach \textbf{KerGM}. While \textbf{mALS} proved to be very robust to noise compared to other methods, the much lower complexity of \textbf{mSync} makes it also a relevant candidate for further studies and extensions to larger populations.

\section{Funding sources}
The data used in the preparation of this article were obtained from OASIS: Cross-Sectional: Principal Investigators: D. Marcus, R, Buckner, J, Csernansky J. Morris; P50 AG05681, P01 AG03991, P01 AG026276, R01 AG021910, P20 MH071616, U24 RR021382.
The project leading to this publication has received funding from Excellence Initiative of Aix-Marseille University - A*MIDEX, a french "Investissements d'Avenir" programme (AMX-19-IET-002).
The research leading to these results has also been supported by the ANR SulcalGRIDS Project, Grant ANR-19-CE45-0014 and the ERA-NET NEURON MULTI-FACT Project, Grant ANR-21-NEU2-0005 funded by the French National Research Agency.


\section{Declaration of competing interest}
The authors declare that they have no known competing financial interests or personal relationships that could have appeared to influence the work reported in this paper.

\section{Author contributions}
We report individual contributions to the paper using the relevant CRediT roles: 

R.Yadav:Conceptualization; Data curation; Formal analysis; Investigation; Methodology; Software; Visualization; Writing - original draft; Writing - review \& editing.

F.-X.Dupé: Conceptualization; Formal analysis; Funding acquisition; Investigation; Methodology; Supervision; Writing - original draft; Writing - review \& editing.

S.Takerkart: Conceptualization; Data curation; Formal analysis; Investigation; Methodology; Software; Supervision; Writing - original draft; Writing - review \& editing.

G.Auzias: Conceptualization; Data curation; Formal analysis; Funding acquisition; Investigation; Methodology; Project administration; Resources; Software; Supervision; Validation; Visualization; Writing - original draft; Writing - review \& editing.

\bibliography{references.bib}

\appendix
\section*{Appendices}

\section{Additional description of the $\beta$-binomial distribution}
\label{annex1}
In the following section we provide additional description of the formulation of the $\beta$-binomial distribution.

The $\beta$-binomial distribution is parameterized by $\nu$, $\alpha$ and $\beta$.
The parameter $\nu$ defines the size of support (in our case, maximum number of $n_{o}$/$n_{s}$).
The setting of $\nu$ can impact the skewness of the distribution but the shape will be Gaussian as long as the value is sufficiently large.
We show in figure \ref{fig:nu_distributions} the $\beta$-binomial distributions for different values of $\nu$, with $\alpha=7.15$ and $\beta=28.62$. In this work, we set $\nu = 30$ which is sufficient to get a distribution close to Gaussian for all combinations of $\alpha$ and $\beta$ parameters that are relevant in our context.

\renewcommand{\thefigure}{A.1}
\begin{figure}[h!]
    \centering
    \includegraphics[width=1\linewidth]{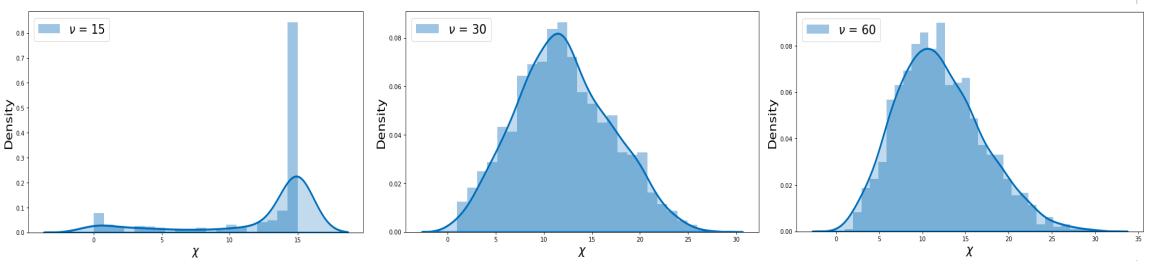}
    \caption{Effect on $\beta$-binomial mass function for different values of $\nu$ fixing $\alpha=7.15$ and $\beta=28.62$}
    \label{fig:nu_distributions}
\end{figure}

\begin{figure}[h!]
    \centering
    \includegraphics[width=0.8\linewidth]{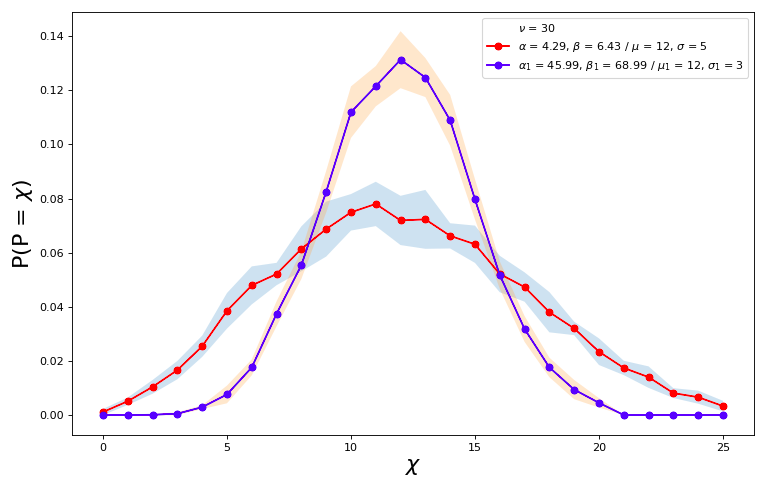}
    \caption{$\beta$-binomial distributions for identical mean: $\mu, \mu_{1} = 12$ but different standard deviations: $\sigma=3, \sigma_1 = 5$. The dotted lines signifies the mean of the distribution where as the shaded area is the standard deviation across 5 trials.}
    
    \label{fig:beta-binom}
\end{figure}

Although $\alpha$ and $\beta$ are not trivial to calibrate, they can be related to $\mu$ and $\sigma$ of a Gaussian distribution using the following formulae:
\begin{equation}
    \rho = \frac{\nu-\mu}{\mu}, \qquad
    \alpha = \frac{(1+\rho)^2  \sigma - n^2  \rho}{ (\nu\rho(1+\rho)-\sigma (1+\rho)^3}, \qquad
     \beta = \frac{\nu-\mu}{\mu}  \alpha 
     \tag{A.1}
     \label{eq:beta_estimate}
\end{equation}

which allows us to control for $\mu$ and $\sigma$, i.e. the amount of nodes to suppress($n_{s}$) and outliers to add($n_{o}$). Figure \ref{fig:beta-binom} illustrates how we can obtain distributions close to two Gaussian with identical $\mu$ but different $\sigma$ value, by controlling $\alpha$ and $\beta$.

\section{Supplementary data showing the fit between simulated and real graphs}
\label{annex2}

As stated in section in \ref{sec:simu_params} we empirically set the simulation parameters $\mu_{pert}, \sigma_{pert}$ and $p = 10\%$ such that our simulated graphs follow the intrinsic properties of real graphs. With the choice of $\mu_{pert}, \sigma_{pert}$ we estimate the corresponding $\alpha$ and $\beta$ of $\beta$-binomial distribution as described in Appendix \ref{annex1}. This allows us to generate graphs with similar mean and standard deviation of number of nodes as in the real data. 
This control on the number of nodes is independent from the other types of perturbations we induce.
In particular, we show on Figure \ref{fig:nb_nodes_distribution} the match between simulated and real graphs for various values of the parameter controlling for the perturbation of the coordinates of the nodes, $\kappa$.
This figure shows the density distribution for the number of nodes in the simulated graphs for different values of $\kappa$, compared to number of nodes in the real population. The largely overlapping distributions confirm the match of the number of nodes, for any value of $\kappa$.
\renewcommand{\thefigure}{B.1}
\begin{figure}[htb]
    \centering
    \includegraphics[width=0.8\linewidth]{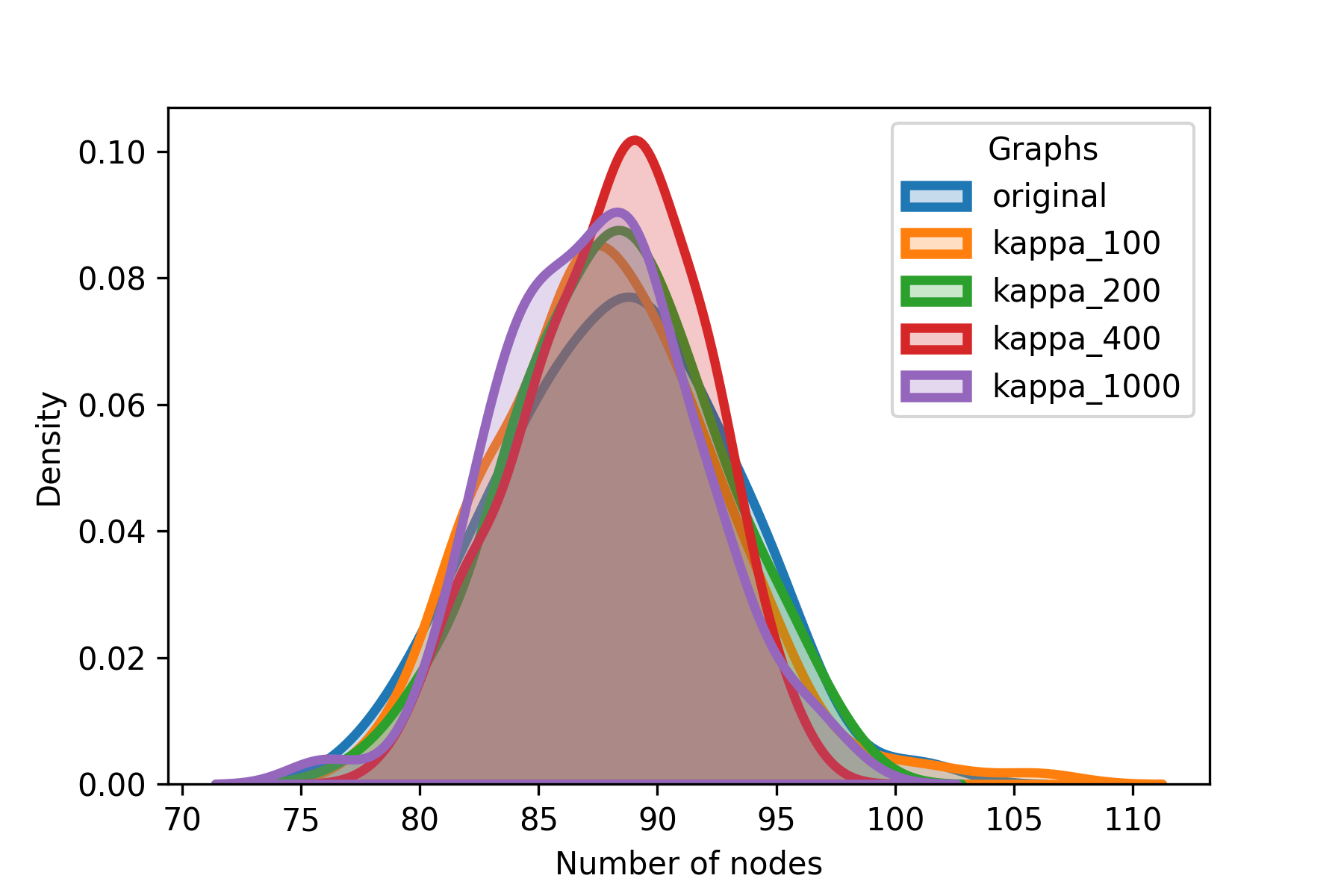}
    \caption{Distribution for number of node in the simulated population corresponding to different $\kappa$ value with the distribution for number of nodes in the real sulcal graphs.}
    \label{fig:nb_nodes_distribution}
\end{figure}

In addition, we also compare the distributions of the geodesic length of the edges which serves as the feature on the edges.
Figure \ref{fig:geo_dist} shows the distributions of mean geodesic distance across a populations of real data and simulated graphs for different values of $\kappa$. The shaded area surrounding each curve shows the standard deviation across a population of 137 graphs in both real and simulated population.
As stated in section \ref{sec:gen_ref} the distance between the nodes in the real graphs are larger than a minimum distance, which is illustrated by the flat portion of the blue curve for low geodesic distances.
Our simulations do not reproduce this feature, as expected from the uniform sampling of the location of outliers nodes that can get close to previous nodes (figure \ref{fig:visu_simu}.b,c).
Note that the fit is good for larger geodesic distance values.

Finally, we show on Figure \ref{fig:degree_dist} the distribution of the degree of nodes for simulated and real graphs. 
The degree corresponds to the number of neighbors of each node, and is thus indicative of the local topology of the graphs.
This figure confirms the good match between simulated and real data, independently of $\kappa$ that controls the perturbation level.

Overall, all our measures confirmed a good match between simulated and real graphs, for any value of $\kappa$.

\renewcommand{\thefigure}{B.2}
\begin{figure}[htb]
    \centering
    \includegraphics[width=0.9\linewidth]{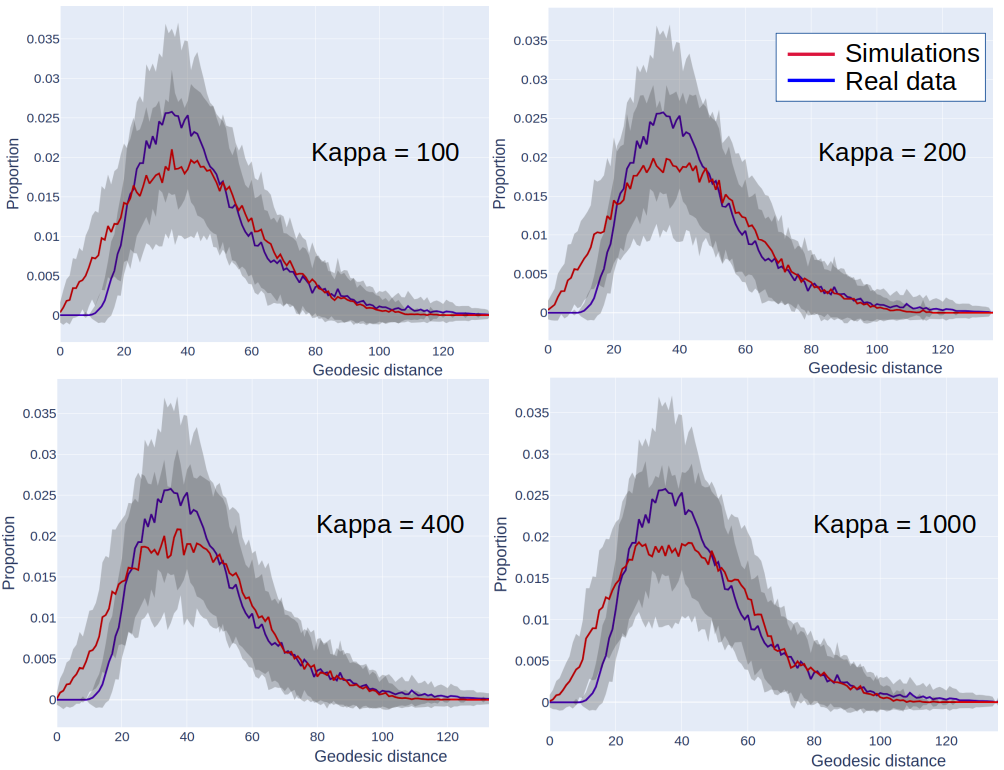}
    \caption{Distribution for geodesic distance in the simulated and real population of 137 graphs. The shaded region corresponds to standard deviations across graphs in the population.}
    \label{fig:geo_dist}
\end{figure}

\renewcommand{\thefigure}{B.3}
\begin{figure}[htb]
    \centering
    \includegraphics[width=0.9\linewidth]{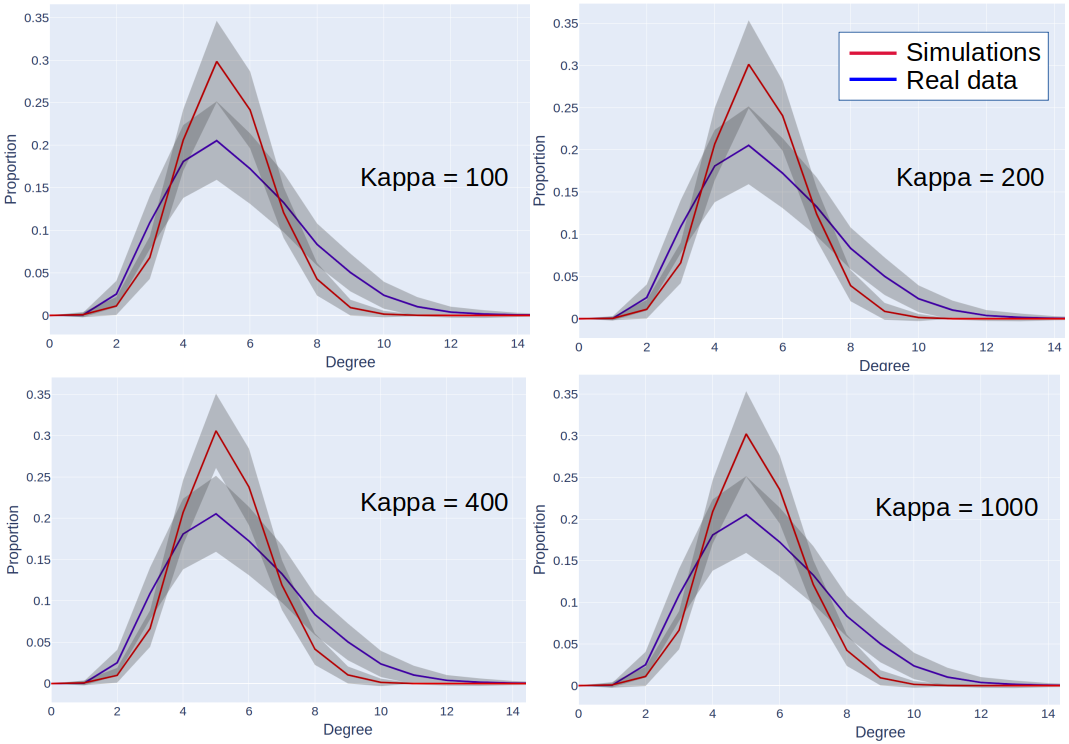}
    \caption{Degree distribution in the simulated and real population of 137 graphs. The shaded region corresponds to standard deviations across graphs in the population.}
    \label{fig:degree_dist}
\end{figure}
\end{document}